\documentclass[review]{elsarticle}

\usepackage{lineno,hyperref}
\modulolinenumbers[5]

\journal{}

\usepackage[cmex10]{amsmath}
\usepackage{mathtools}
\usepackage{graphicx}
\usepackage{subfig}
\usepackage{cite}
\usepackage{epstopdf}
\usepackage{fixltx2e}
\usepackage{amsfonts}
\usepackage{multirow}
\usepackage{tabularx}
\usepackage{array}
\usepackage[tableposition=top]{caption}
\usepackage{color,soul}
\usepackage{longtable}
\usepackage{caption}
\usepackage{amsthm}
\theoremstyle{definition}
\newtheorem{definition}{Definition}
\newtheorem{theorem}{Theorem}
\newtheorem{corollary}{Corollary}
\newtheorem{lemma}{Lemma}

\usepackage{tikz}
\usetikzlibrary{calc}
\usepackage{mfirstuc}
\usepackage{glossaries}
\usepackage{paralist}
\usepackage{enumitem}
\setenumerate{noitemsep}
\begin{document}
\tikzstyle{every picture}+=[remember picture]
\begin{frontmatter}

\title{Ground Truth Bias in External Cluster Validity Indices}

\author[mymainaddress]{Yang Lei\corref{mycorrespondingauthor}} 
\cortext[mycorrespondingauthor]{Corresponding author}
\ead{yalei@student.unimelb.edu.au}

\author[mymainaddress]{James C. Bezdek}
\ead{jbezdek@unimelb.edu.au}
\author[mymainaddress]{Simone Romano}
\ead{simone.romano@unimelb.edu.au}
\author[mymainaddress]{Nguyen Xuan Vinh}
\ead{vinh.nguyen@unimelb.edu.au}
\author[mysecondaryaddress]{Jeffrey Chan}
\ead{jeffrey.chan@rmit.edu.au}
\author[mymainaddress]{James Bailey}
\ead{baileyj@unimelb.edu.au}

%\ead{\{jbezdek, simone.romano, vinh,nguyen, baileyj\}@unimelb.edu.au}
\address[mymainaddress]{Department of Computing and Information Systems \\ The University of Melbourne, Victoria, Australia}
\address[mysecondaryaddress]{School of Science (Computer Science and Information Technology) \\ RMIT University, Victoria, Australia}



%The author names and affiliations could be formatted in two ways:
%\begin{enumerate}[(1)]
%\item Group the authors per affiliation.
%\item Use footnotes to indicate the affiliations.
%\end{enumerate}
%See the front matter of this document for examples. You are recommended to conform your choice to the journal you are submitting to.
\begin{abstract}
External \emph{cluster validity indices} (CVIs) are used to quantify the quality of a clustering by comparing the similarity between the clustering and a ground truth partition. However, some external CVIs show a biased behaviour when selecting the most similar clustering. Users may consequently be misguided by such results. Recognizing and understanding the bias behaviour of CVIs is therefore crucial.
 
It has been noticed that some external CVIs exhibit a preferential bias towards a  larger or smaller number of clusters which is monotonic (directly or inversely) in the number of clusters in candidate partitions. This type of bias is caused by the functional form of the CVI model. For example, the popular \emph{Rand Index} (RI) exhibits a monotone increasing (NCinc) bias, while the \emph{Jaccard Index} (JI) index suffers from a monotone decreasing (NCdec) bias. This type of bias has been previously recognized in the literature.

In this work, we identify a new type of bias arising from the distribution of the ground truth (reference) partition against which candidate partitions are compared. We call this new type of bias \emph{ground truth} (GT) bias. This type of bias occurs if a change in the reference partition causes a change in the bias status (e.g., NCinc, NCdec) of a CVI. For example, NCinc bias in the RI can be changed to NCdec bias by skewing the distribution of clusters in the ground truth partition. It is important for users to be aware of this new type of biased behaviour, since it may affect the interpretations of CVI results.

The objective of this article is to study the empirical and theoretical implications of GT bias. To the best of our knowledge, this is the first extensive study of such a property for external cluster validity indices. Our computational experiments show that $5$ of $26$ indices studied in this paper exhibit GT bias. Following the numerical examples, we provide a theoretical analysis of GT bias based on the relationship between the RI and quadratic entropy. Specifically, we prove that the quadratic entropy of the ground truth partition provides a computable test which predicts the NC bias status of the Rand Index.
\end{abstract}

\begin{keyword}
External Cluster Validity Indices\sep Rand Index\sep Ground Truth Bias\sep Quadratic Entropy
\end{keyword}

\end{frontmatter}

\linenumbers

\section{Introduction}
Clustering is one of the fundamental techniques in data mining, which helps users explore potentially interesting patterns in unlabeled data. Cluster analysis has been widely used in many areas, ranging from bioinformatics~\citep{bioinfor2007cluster} and market segmentation~\citep{market1983cluster} to information retrieval~\citep{ir2013cluster} and image processing~\citep{image2006cluster}. However, depending on different factors, e.g., different clustering algorithms, initializations, parameter settings (the number of clusters $c$), many alternative candidate partitions might be discovered for a fixed dataset.

Cluster validity measures are used to quantify the goodness of a partition. 
Many CVIs have been proposed and successfully used for this task~\citep{vinh2010,jain1988}. These measures can be generally divided into two types: internal and external. If the data are \emph{labeled}, the ground truth partition can be used with an external CVI to explore the match between candidate and ground truth partitions. Since the labeled data may not correspond to clusters proposed by any algorithm, we will refer groups in the ground truth as \emph{subsets}, and algorithmically proposed groups as \emph{clusters}. When the data are \emph{unlabeled} (the real case), an important post-clustering question is how to evaluate different candidate partitions. This job falls to the internal CVIs. One of the most important uses of the external CVIs is to evaluate the comparative quality of internal CVIs on labeled data~\citep{internalcompare2013}, so that in the real case, some confidence can be placed in a chosen internal CVI to guide us towards realistic clusters found in unlabeled data. \emph{This article is focused on external CVIs}.

External CVIs (or comparison measures), are often interpreted as similarity (or dissimilarity) measures between the ground truth and candidate partitions. The ground truth partition, which is usually generated by an expert in the data domain, identifies the primary substructure of interest to the expert. This partition provides a benchmark for comparison with candidate partitions. The general idea of this evaluation methodology is that the more similar a candidate is to the ground truth (a larger value for the similarity measure), the better this partition approximates the labeled structure in the data.  

However, this evaluation methodology  implicitly assumes that the similarity measure works correctly, i.e., that a larger similarity score indicates a partition that is really more similar to the ground truth.
But this assumption may not always hold. When this assumption is false, the evaluation results will be misleading. One of the reasons that can cause the assumption to be false is that a measure may have bias issues. That is, some measures are biased towards certain clusterings, even though they are not more similar to the ground truth compared to the other  candidate partitions being evaluated. This can cause misleading results for users employing these biased measures. Thus, recognizing and understanding the bias behaviour of the CVIs is crucial.

The \emph{Rand Index} (RI, similarity measure) is a very popular pair-counting based validation measure that has been widely used in many applications~\citep{RI2010nature, RI2011a, RI2012b, RI2013a, RI2014a, rivi2015} in the last five years. It has been noticed that the RI tends to favor candidate partitions with larger numbers of clusters when the number of subsets in the ground truth is fixed~\citep{vinh2010}, i.e., it tends to increase as the number of clusters increases (we call it NCinc bias in this work, where NC $=$ number of clusters).
NC bias means that the CVI's preference is influenced by the number of clusters in the candidate partitions. For example, some measures may prefer the partition with larger (smaller) number of clusters, i.e., NCinc (NCdec) bias.
The following initial example illustrates NC bias for two popular measures, the \emph{Rand Index} (RI) and \emph{Jaccard Index} (JI) measures. 

\subsection{\textbf{Example 1 - NC bias of RI and JI}\label{subsec:exp1}} 
In this example, we illustrate NC bias for RI and JI. We generate a set of candidate partitions randomly with different numbers of clusters and a random ground truth. We use RI and JI to choose the most similar partition from the candidate partitions by comparing the similarity between each of them and the ground truth. As there is no difference in the generation methodology of the candidate partitions, we expect them to be treated equally on average. A measure without NC bias should treat these candidate partitions equally without preference to any partition in terms of their different number of clusters. However, if a measure prefers the partition, e.g., with a larger number of clusters (gives higher value to the partition with a larger number of clusters if it is a similarity measure), we say it possess NC bias, more specifically, NCinc bias. 

\begin{figure}[t!]
\centering
\subfloat[Average RI values with random $U_{GT}$ containing $5$ subsets.]{\includegraphics[width = 0.4\textwidth]{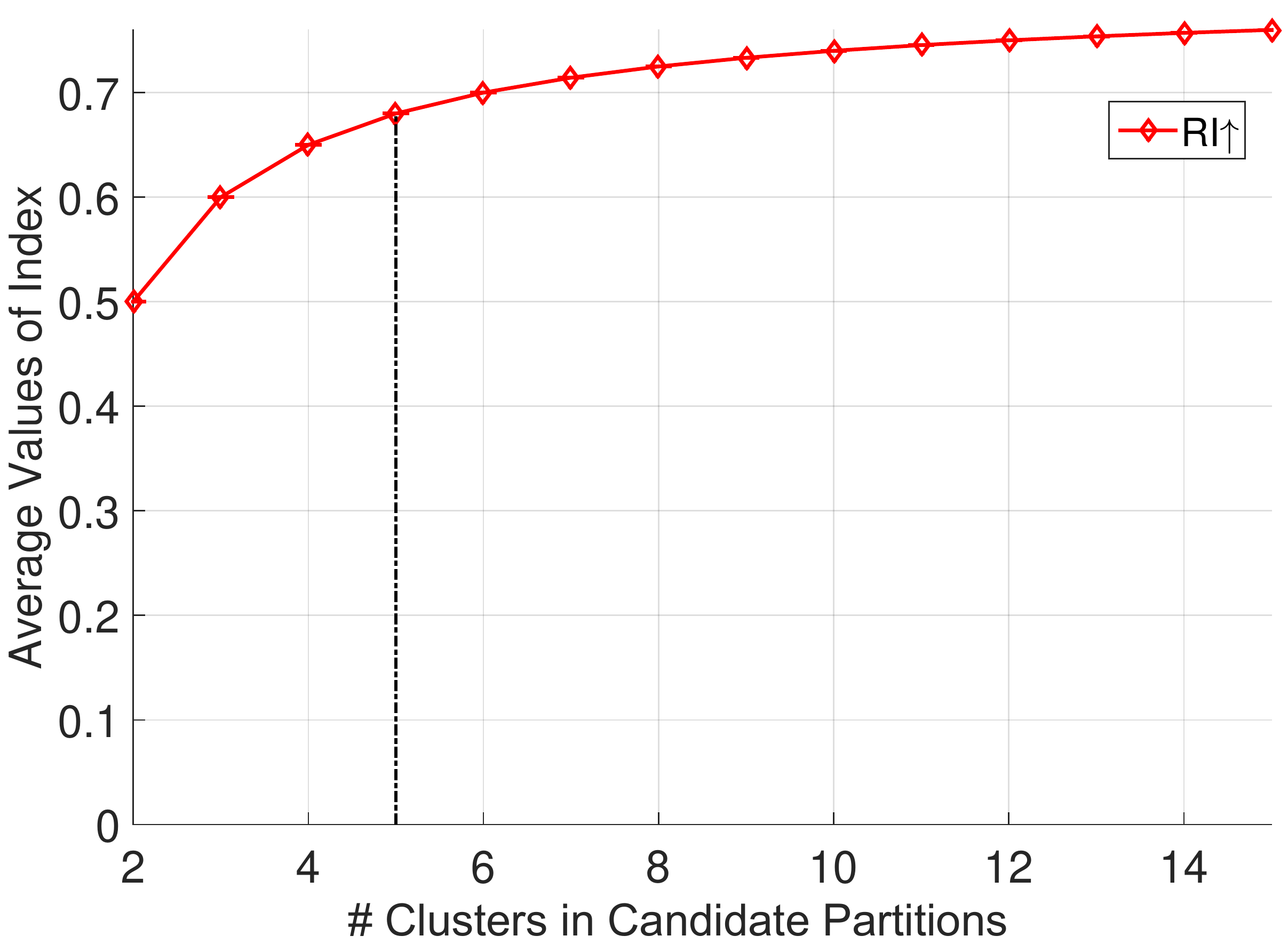}
\label{subfig:exUniGTRI}}
\qquad
\subfloat[Averge JI values with random $U_{GT}$ containing $5$ subsets.]{\includegraphics[width = 0.4\textwidth]{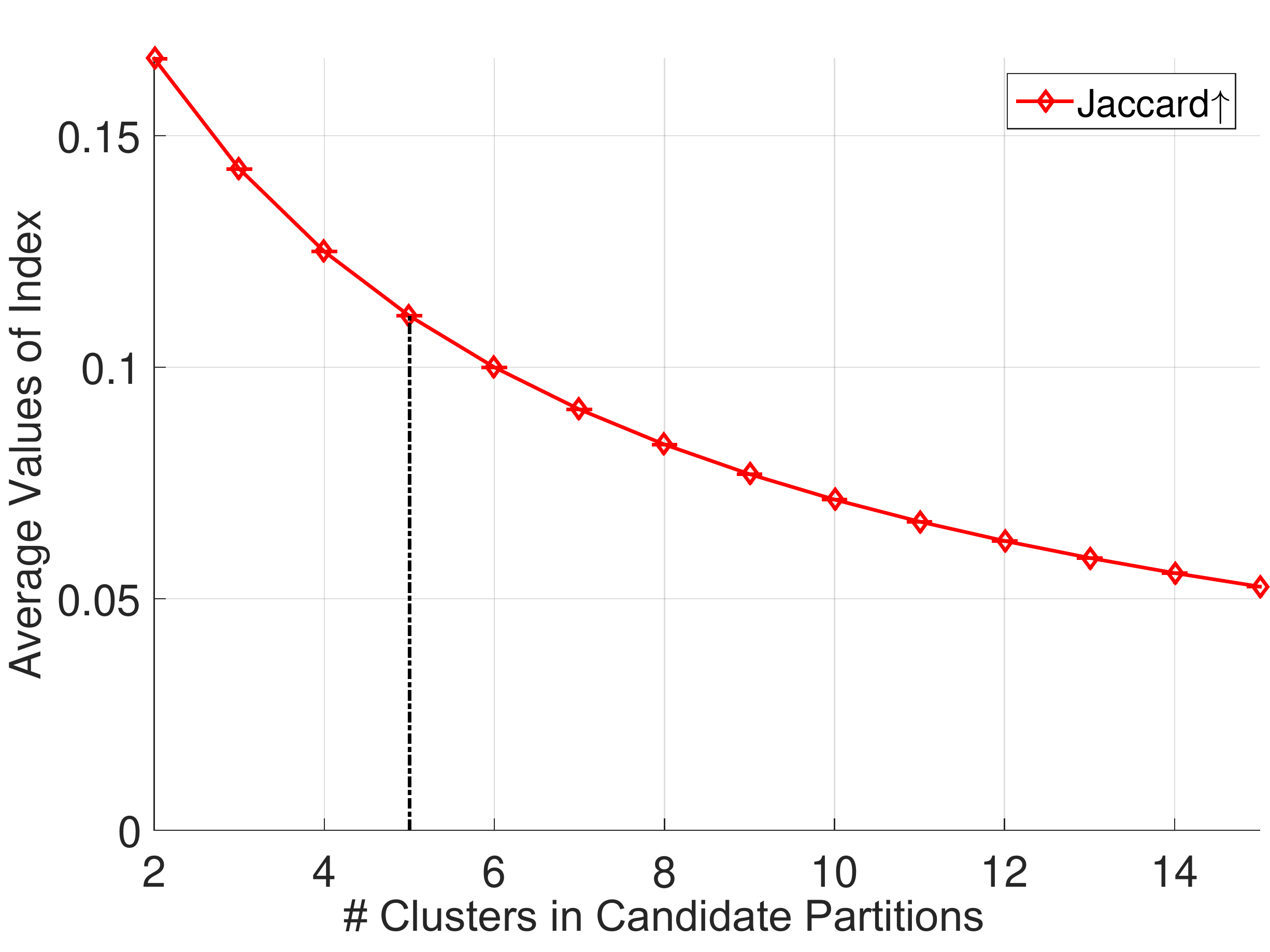}
\label{subfig:exUniGTVI}}
\caption{The average RI and JI values over $100$ partitions at each $c$ with uniformly generated ground truth. The symbol $\uparrow$ means larger values are preferred. Vertical line indicates correct number of clusters.}
\label{fig:exUniGT}
\end{figure}

Let $U_{GT}$ be a ground truth partition with $c_{true}$ subsets. 
Consider a set of $N=100,000$ objects, let the number of clusters in the candidate partitions $c$ vary from $2$ to $c_{max}$, where $c_{max} = 3* c_{true}$. We randomly generate a ground truth partition $U_{GT}$ with $c_{true}=5$. Then for each $c$, $2\leq c \leq 15$, we generate $100$ partitions randomly, and calculate the RI and JI between $U_{GT}$ and each generated partition. Finally, we compute the average values of these two measures at each value of $c$.   The results are shown in Figure~\ref{fig:exUniGT}. 
Please note that the RI and JI are max-optimal (larger value is preferred). Evidently RI monotonically increases and JI monotonically decreases as $c$ increases. Figure~\ref{fig:exUniGT} shows that for this experiment, the RI points to $c=15$, its maximum over the range of $c$; and the JI points to $c=2$, its maximum over the range of $c$.
Both indices exhibit NC bias (RI shows NCinc bias and JI shows NCdec bias).

\begin{figure}[t!]
\centering
\subfloat[Average RI values with skewed ground truth. ]{\includegraphics[width = 0.4\textwidth]{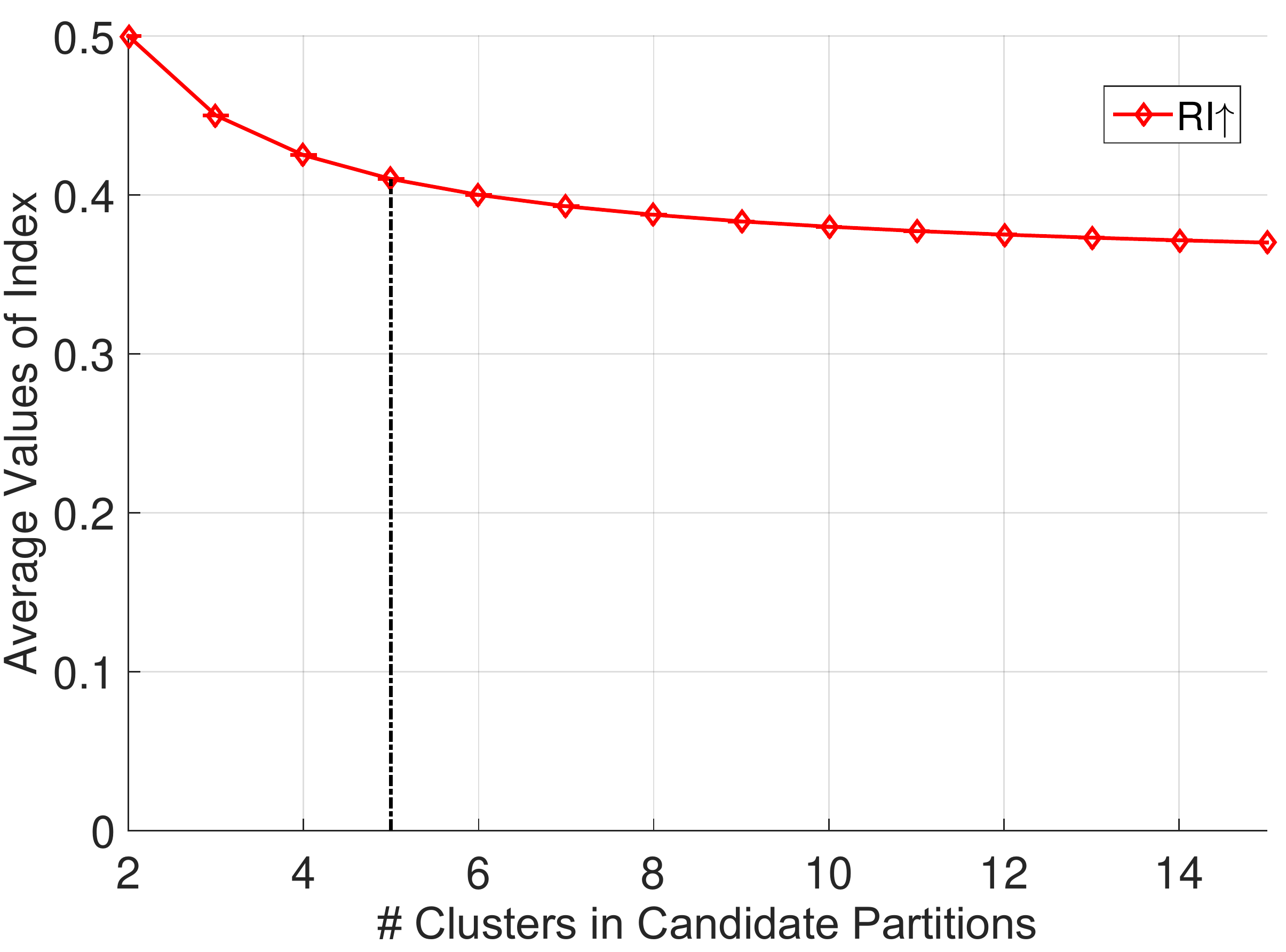}
\label{subfig:exSkewGTRI}}
\qquad
\subfloat[Averge JI values with skewed ground truth.]{\includegraphics[width = 0.4\textwidth]{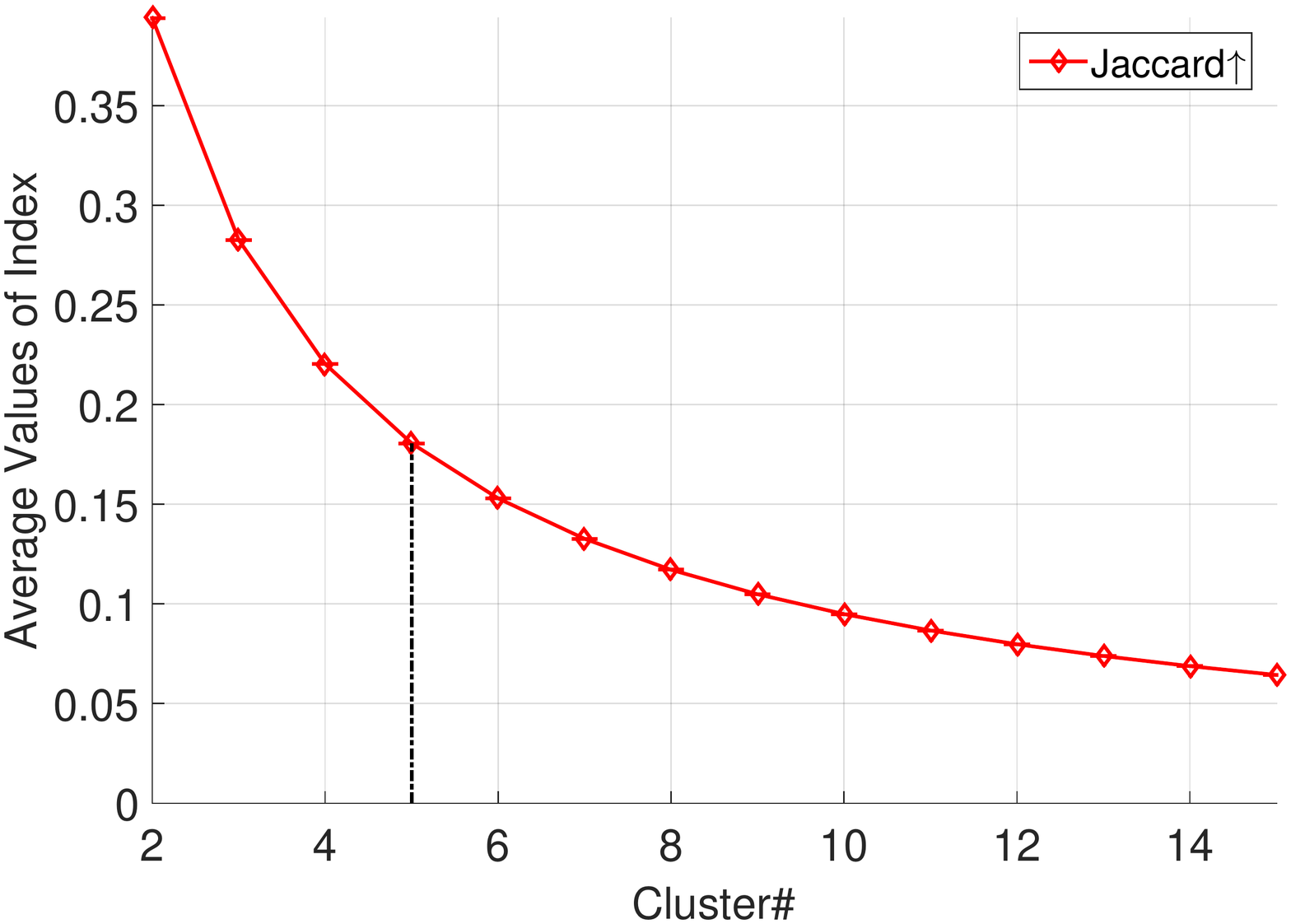}
\label{subfig:exSkewGTVI}}
\caption{The average RI and JI values over $100$ partitions at each $c$ with skewed ground truth. The symbol $\uparrow$ means larger values are preferred. Vertical line indicates correct number of clusters.}
\label{fig:exSkewGT}
\end{figure}
 
But, does the RI \emph{always} exhibit NCinc bias towards clusterings with a larger numbers of clusters? \emph{The answer is no}. We have discovered that the overall bias of some CVIs, including the RI, may change their NC bias tendencies depending on the distribution of the subsets in the ground truth. The change in the NC bias status of an external CVI due to the different ground truths  is called \emph{GT bias}. This kind of changeable bias behaviour caused by the ground truth has not been recognized previously in the literature. It is important to be aware of this phenomenon, since it affects how a user should interpret clustering validation results.
Next, we give an example of GT bias (GT $=$ ground truth).

\subsection{\textbf{Example 2 - GT bias of RI}\label{subsec:exp2}} 
We use the same protocols as in Example $1$, but change the distribution of the subsets in the ground truth by randomly assigning $80\%$ of the objects to the first cluster and then randomly assigning the remaining $20\%$ of the labels to the other four clusters for $c=2,3,4,5$. Thus, the distribution of the ground truth is  heavily skewed (non-uniform). The average values of RI and JI are shown in Figure~\ref{fig:exSkewGT}. The shape of JI in Figures~\ref{subfig:exUniGTVI} and~\ref{subfig:exSkewGTVI} is same:  it still decreases monotonically with $c$, exhibiting NCdec bias, and indicating $c=2$ as its preferred choice. 
Turning now to the RI, we see that trend seen in Figure~\ref{subfig:exUniGTRI} is reversed. The RI in Figure~\ref{subfig:exSkewGTRI} is maximum at $c=2$, and \emph{decreases} monotonically as $c$ increases. So the NC bias of RI has changed from NCinc bias to NCdec bias. Thus, RI shows GT bias. 
To summarize, Examples $1$ and $2$ show that NC bias is possessed by some external CVIs due to monotonic tendencies of the underlying mathematical model. But beyond this, some external CVIs can be influenced by GT bias, which is due to the way the distribution of the ground truth interacts with the elements of the CVI.
 
The objective of this article is to study the empirical and theoretical implications of GT bias. To the best of our knowledge, this is the first extensive study of this property for external cluster validity indices. In this work, our contributions can be summarized as follows:
\begin{enumerate}
 \item We identify the GT bias effect for external validation measures, and also explain its importance.
\item We test and discuss NC bias for $26$ popular pair-counting based  external validation measures. 
 \item We prove that RI and related $4$ indices suffer from GT bias. And also provide theoretical explanations for understanding why GT bias happens and when it happens on RI and related $4$ indices.
 \item We present experimental results that support our analysis. 
 \item We present an empirical example to show that \emph{Adjusted Rand index} (ARI) also suffers from a modified GT bias.
\end{enumerate}
The remainder of the paper is organized as follows. In Section~\ref{sec:Rwork} we discuss work related to the bias problems of some external validation measures. We introduce relevant notations and definitions of NC bias and GT bias in Section~\ref{sec:notdef}. In Section~\ref{sec:bag}, we briefly introduce some background knowledge about $26$ pair-counting based external validation measures. In section~\ref{sec:test}, we test the influence of NC bias and GT bias for these $26$ measures. Theoretical analysis of GT bias on the RI is presented in Section~\ref{sec:anly}.  An experimental example, showing that ARI has GT bias in certain scenarios, is presented in Section~\ref{sec:ari}. The paper is concluded in Section~\ref{sec:cons}. 

\section{Related Work} \label{sec:Rwork}
Several works have discussed the bias behaviour of external CVIs. As the conditions imposed on the discussion of the biased behaviour are varied, here we classify these conditions into three categories for convenience of discussion:
\begin{inparaenum}[i)]
\item general bias;
\item NC bias;
\item GT bias.
\end{inparaenum}
\paragraph{\textbf{General Bias}}
It has been noticed that the RI exhibits a monotonic trend as both the number of subsets in the ground truth and the number of clusters in the candidate partitions increases~\citep{fowlkes1983,relative2010,means2010}. However, in our case, we consider the monotonic bias behaviour of an external CVI as a function of the number of clusters in the candidate partitions when the number of subsets in the ground truth is fixed. 

Wu et al.~\citep{kmeansuniform2009} observed that some external CVIs were unduly influenced by the well known tendency of k-means to equalize cluster sizes. They noted that certain CVIs tended to prefer approximately balanced k-means solutions even though the ground truth distribution was heavily skewed. 
The only case considered in~\citep{kmeansuniform2009} was the special case when all of the candidate partitions had the same number of clusters. We will develop the general case, allowing candidate partitions to have different numbers of clusters. 

Wu et al.~\citep{fmeasure2010} studied the use of the external CVI known as the F-measure for evaluation of clusters in the context of document retrieval. They found that the F-measure tends to assign higher scores to partitions containing a large number of clusters, which they called the ``the incremental effect'' of the F-measure. 
These authors also found that the F-measure has a ``prior-probability effect'', i.e., the F-measure tends to assign higher scores to partitions with higher prior probabilities for the relevant documents. Wu et al. only discussed using the F-measure for accepting or rejecting proposed documents, they did not consider the multiclass case. 
\paragraph{\textbf{NC Bias}} 
The NC bias problem of some external CVIs has been noticed in the literature~\citep{milligan1986, vinh2010, simone2014}. Nguyen et al.~\citep{vinh2010} pointed out that some external validation measures such as the \emph{mutual information} (MI) (also the work~\citep{simone2014})  and the \emph{normalized mutual information} (NMI) suffered from NCinc bias. Based this observation, they proposed adjustments to the information-theoretic based measures. However, they did not notice that the CVIs may show different NC bias behaviour with different ground truth partitions.
\paragraph{\textbf{GT Bias}}
Milligan and Cooper~\citep{milligan1986} tested $5$ external CVIs, i.e., RI,  \emph{Adjusted Rand Index} (ARI, Hubert \& Arabie)~\citep{arabie1973}, ARI (Morey \& Agresti)~\citep{morey1984}, Fowlkes \& Mallow (FM)~\citep{fowlkes1983} and \emph{Jaccard Index} (JI), by comparing partitions with variable numbers of clusters generated by the hierarchical clustering algorithms, against the ground truth. The empirical tests showed that the RI suffered from NCinc bias, and FM and JI suffered from NCdec bias. However, it was mentioned in this work that ``... the bias with the Rand index would be to select a solution with a larger number of clusters. The only exception occurred when two clusters were hypothesized to be present in the data. In this case, the bias was reversed.'' This empirical observation can be related our work. However, there was no analysis or further discussion about this reversed bias behaviour of RI except this isolated observation. In this work, we provide a comprehensive empirical and theoretical study of this kind of changeable bias behaviour due to the distribution of the ground truth.

\section{Notation and Definitions} \label{sec:notdef}
In this section, we first introduce the notations used in this work. Then we provide the definitions about the different bias behaviours, i.e., NC bias, GT bias which further has two subtypes, i.e., GT1 bias and GT2 bias.
\subsection{Notation}
 
Let $S$ be a set of $N$ objects $\{o_1, \ldots, o_N\}$. A convenient way to represent a crisp $c-partition$ of $S$ is with a set of ($cN$) values $\{u_{ik}\}$ arrayed as a $c \times N$ matrix $U=[u_{ik}]$. Element $u_{ik}$ is the \emph{membership} of $o_k$ in cluster $i$. We denote the set of all possible $c$-partitions of $S$ as:
\begin{equation}
M_{hcN} = \{U \in \mathfrak{R}^{cN}| \forall i, k, u_{ik} \in \{0,1\} ; \forall k, \sum_{i=1}^c u_{ik}=1; \forall i, \sum_{k=1}^N u_{ik} >0 \}
\end{equation}

The cardinality (or size) of cluster $i$ is $\sum_{k=1}^N u_{ik} = n_i$. When all of the $n_i$ are equal to $N/c$, we say that $U$ is \emph{balanced}. 

\subsection{Definitions}
This section contains definitions for the types of bias exerted on external CVIs by their functional forms (NC bias) and the distribution of the ground truth partition (GT bias). 
We will call the influence of the number of clusters in ground truth partition, $U_{GT}$, Type 1 or \emph{GT1 bias}, and the influence of the size distribution of the subsets in $U_{GT}$ Type 2, or \emph{GT2 bias}. 
\begin{definition} \label{def:ncbias}
Let $U_{GT} \in M_{hrN}$ be any crisp ground truth partition with $r$ subsets, where $2 \leq r \leq N$. Let $CP=\{V_1, \ldots, V_m\}$, where $V_i \in M_{hc_iN}$,  be a set of candidate partitions with different numbers of clusters, where $2 \leq c_i \leq N$. We compare $U_{GT}$ with each $V_i \in CP$ using an external \emph{Cluster Validity Index} (CVI) and choose the one that is the best match to $U_{GT}$. There are two types of external CVIs: max-optimal (larger value is better) similarity measures such as \emph{Rand's index} (RI); and min-optimal (smaller value is better) dissimilarity measures such as the Mirkin metric (refer to Table~\ref{tb:RIformu}). \\
We say an external (CVI) has NC bias if it shows bias behaviour with respect to the \emph{number of clusters} in $V_i$ when comparing $V_i \in CP$ to  the ground truth $U_{GT}$. There are three types of NC bias:
\begin{enumerate}
\item if a max-optimal (min-optimal) CVI tends to assign higher (smaller) scores to the partition $V_i \in CP$ with larger $c_i$, then we say this CVI has NCinc (NC increase) bias;
\item if a max-optimal (min-optimal) CVI tends to assign smaller (higher) scores to the partition $V_i \in CP$ with larger values of $c_i$, then we say this CVI has NCdec (NC decrease) bias;
\item if a CVI tends to be indifferent to the values of $c_i$ for the partitions $V_i \in CP$, we say that this CVI has no NC bias, i.e., NCneu (NC neutral) bias.
\end{enumerate}
\end{definition}

Next, we define \emph{ground truth} bias (GT bias), which occurs if the use of a \emph{different ground truth partition} alters the NC bias status of an external CVI.
 
\begin{definition}\label{def:gtbias}
Let $Q$ and $Q^\prime$ denote the NC bias status of an external CVI, $\mathcal{CVI}$, with respect to two ground truth partitions, $U_{GT}$ and $U_{GT}^\prime$ respectively, so $Q, Q^\prime \in \{\text{NCinc}, \text{NCdec}, \text{NCneu}\}$. If $Q \neq Q^\prime$, then $\mathcal{CVI}$ has \emph{ground truth bias} (GT bias).
\end{definition}
For example, given $U_{GT} \neq U_{GT}^\prime$, if a $\mathcal{CVI}$ shows e.g., NCinc bias with $U_{GT}$, and shows, e.g., NCneu bias with $U_{GT}^\prime$, then this $\mathcal{CVI}$ has GT bias.
Definition~\ref{def:gtbias} characterizes GT bias as an transition effect on the NC bias status of  $\mathcal{CVI}$. There are quite a few subcases of $GT$ bias depending on the properties of $U_{GT}$ and $U_{GT}^\prime$ relative to each other. 
In this article we have studied two specific cases in GT bias, i.e., GT1 bias and GT2 bias. Generally speaking, one external CVI changes its bias status with two ground truth $U_{GT1}$ and $U_{GT2}$: 
\begin{inparaenum}[i)]
\item GT1 bias,  the subsets in these two ground truths are uniformly distributed but with different numbers of subsets; 
\item GT2 bias, these two ground truths have same number of subsets but with different distributions. The formal definitions of GT1 bias and GT2 bias are described as follows.
\end{inparaenum}
\begin{definition} \label{def:gt1bias}
Let $U_{GT} \in M_{hrN}$ be a balanced crisp ground truth partition with $r$ subsets $\{u_1, \ldots, u_r\}$, i.e., $p_i = \frac{|u_i|}{N} = \frac{1}{r}$ , and $U_{GT}^\prime \in M_{hr^{\prime}N} $ be a balanced crisp ground truth partition with $r^\prime$ subsets $\{u_1^\prime, \ldots, u_{r^{\prime}}^\prime\}$, i.e., $p_i^\prime = \frac{|u_i^\prime|}{N} = \frac{1}{r^\prime}$, where $\mathbf{r \neq r^\prime}$. 
We say an external CVI has \textbf{{GT1 bias}} if the NC bias status for $U_{GT}$ is different from that of  $U_{GT}^\prime$.
\end{definition}

For example, given $U_{GT}$ with $2$ balanced subsets, and $U_{GT}^\prime$ with $5$ balanced subsets, then if an CVI shows e.g., NCneu bias with $U_{GT}$, and NCinc bias with $U_{GT}^\prime$, then this CVI has GT1 bias.

\begin{definition} \label{def:gt2bias}
Let $U_{GT} \in M_{hrN}$ be a crisp ground truth partition with $r$ subsets $\{u_1, \ldots, u_r\}$, $P = \{p_1, \ldots, p_r\}=\{\frac{|u_1|}{N}, \ldots, \frac{|u_r|}{N}\}$ and $p_2 = p_3 = \ldots = p_r = \frac{1-p_1}{r-1}$.  Let $U_{GT}^\prime \in M_{hr^{\prime}N}$ be another crisp ground truth partition with $r^\prime$ subsets $\{u_1^\prime, \ldots, u_{r^{\prime}}^\prime\}$ and $P^\prime = \{p_1^\prime, \ldots, p_{r^\prime}^\prime\}= \{\frac{|u_1^\prime|}{N}, \ldots, \frac{|u_{r^\prime}^\prime|}{N}\}$, $p_2^\prime = p_3^\prime = \ldots = p_{r^\prime}^\prime = \frac{1-p_1^\prime}{r-1}$, where $\mathbf{r= r\prime}$ and $\mathbf{p_1 \neq p_1^\prime}$. 
We say an external CVI has \textbf{\emph{GT2 bias}} if it exhibits different types of NC bias for $U_{GT}$ and $U_{GT}^\prime$.
\end{definition}
For example, given $U_{GT} \in M_{h5N}$ with $p_1 = 0.2$ and $U_{GT}^\prime \in M_{h5N}$ with $p_1^\prime = 0.8$, if an external CVI shows, e.g., NCinc bias for $U_{GT}$ and shows e.g., NCdec bias for $U_{GT}^\prime$, then this CVI has GT2 bias.

\begin{figure}[ht!]
\centering
\includegraphics[width=0.8\textwidth]{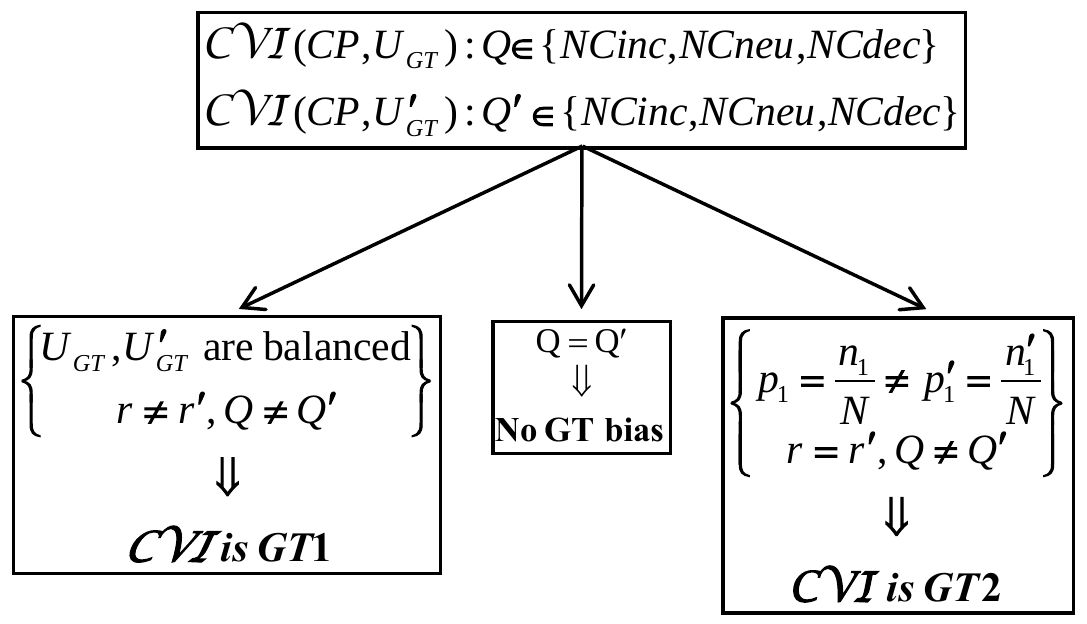}
\caption{The relationship between NC bias and GT bias in Definitions~\ref{def:ncbias}-~\ref{def:gt2bias}.}
\label{fig:gtbias}
\end{figure}

Figure~\ref{fig:gtbias} illustrates the relationship between NC bias and GT bias that is contained in Definitions~\ref{def:ncbias} -~\ref{def:gt2bias}. In this Figure, $CP$ denotes a set of crisp candidate partitions with different numbers of clusters, and $\mathcal{CVI}$ denotes an external CVI. $U_{GT} \neq U_{GT}^\prime$ are different crisp ground truth partitions and $U_{GT} \in M_{hrN}, U_{GT} \in M_{hr{^\prime}N}$. We summarized the different bias problems discussed in this work in Table~\ref{tb:glory}.

\begin{table} [t!]
\caption{Glossaries about different bias discussed in this paper.}
\renewcommand{\arraystretch}{0.7}
\begin{tabular}{|p{4cm} |p{9cm}|}
\hline
Glossary & Explanation\\
\hline
NC bias & An external CVI shows bias behaviour with respect to the number of clusters in the compared clusterings.\\
\hline
NCinc bias (NC increase) & One of the NC bias status. An external CVI prefers clusterings with larger number of clusters. \\
\hline 
NCdec bias (NC decrease)& One of the NC bias status. An external CVI prefers clusterings with smaller number of clusters.\\
\hline 
NCneu bias (NC neutral)& One of the NC bias status. An external CVI has no bias for clusterings with respect to the number of clusters.\\
\hline
GT bias & An external CVI shows different NC bias status when varying the ground truth.\\
\hline
GT1 bias & A subtype of GT bias. An external CVI shows different NC bias status for two ground truths with uniform distribution but with different numbers of subsets. \\
\hline
GT2 bias & A subtype of GT bias. An external CVI shows different NC bias status for two ground truths that have the same number of subsets but with different subset distributions.\\
\hline
\end{tabular}
\label{tb:glory}
\end{table}

\section{Pair-counting External Cluster Validity Measures} \label{sec:bag}
In this part, we will give some background knowledge briefly about the pair-counting based cluster validity measures. In addition, we also provide a list of $26$ pair-counting based measures which will be tested  for their NC bias and GT bias problems.
\begin{table}[ht!]
\caption{Contingency table based on partitions $U$ and $V$, $n_{ij} = |u_i \cap v_j|$}
\label{tb:contingency} 
\renewcommand{\arraystretch}{0.9}
\centering
\scalebox{0.9}{
	\begin{tabular}{|c| c| c c c  c| c|}
	\hline
		&          & \multicolumn{4}{c|}{$V \in M_{hcN}$}                                                    &  \\
		\hline
		& Cluster & $\mathbf{v_1}$ & $\mathbf{v_2}$ & $\ldots$ & $\mathbf{v_c}$  & Sums \\
		\hline
\begin{tabular}{p{1.8cm}}
		$U \in M_{hrN}$ 
\end{tabular}  & \multicolumn{1}{c|}{
		                        $\begin{matrix}
								\mathbf{u}_1\\
								\mathbf{u}_2 \\
								\vdots \\
								\mathbf{u}_r
 								\end{matrix}$ }
 								& \multicolumn{4}{c|}{$\begin{matrix}  
																	          n_{11} & n_{12} & \ldots & n_{1c} \\
																	          n_{21} & n_{22} & \ldots & n_{2c} \\
																	          \vdots & \vdots &  & \vdots \\
																	          n_{r1} & n_{r2} & \ldots & n_{rc}  
																				\end{matrix}$ }  & \multicolumn{1}{c|}{$\begin{matrix}					                                                                                                                  
																																a_1\\
																																a_2 \\
																																\vdots \\
																																a_r \end{matrix}			
																															$}\\
\hline
	 & Sums  & $b_1$ & $b_2$ & $\ldots$ & $b_c$  & $\sum_{ij} n_{ij} = N$\\ 
\hline
	\end{tabular}
	}
\end{table}

Pair-counting based comparison CVIs are a group of popular measures based on counting the agreements and disagreements between two crisp partitions in terms of shared pairs of objects.
As in Example $1$, we denote the subsets corresponding to the clusters in $U$ and $V$ as $\{u_1, \ldots, u_r\}$ and $\{v_1, \ldots, v_c\}$. Suppose $U \in M_{hrN}$ and $V \in M_{hcN}$ are partitions of $S$. The contingency table that pairs these two partitions is shown in Table~\ref{tb:contingency}. Note that the numbers of clusters in $U$ and $V$ need not be equal, $r\neq
c$.

The entry $n_{ij}$ indicates the number of shared object pairs in clusters $u_i$ and $v_j$. The row sum, $a_i$, is the number of objects in cluster $u_i$ and the column sum, $b_j$,  is the number of objects in cluster $v_j$. 
The number of pairs of shared objects between $U$ and $V$ is divided into four groups: $k_{11}$, the number of pairs that are in the same cluster in both $U$ and $V$; $k_{00}$, the number of pairs that are in different clusters in both $U$ and $V$; $k_{10}$, the number of pairs that are in the same cluster in $U$ but in different clusters in $V$; and $k_{01}$, the number of pairs that are in different clusters in $U$ but in the same clusters in $V$. 
%These four types of pairs can be computed as:
%\begin{align*}
%k_{11} & = \frac{1}{2} \sum_{i=1}^r \sum_{j=1}^c n_{ij} (n_{ij} -1) \\
%k_{00} & = \frac{1}{2} \big( N^2 + \sum_{i=1}^r \sum_{j=1}^c n_{ij}^2 - (\sum_{i=1}^r a_i^2 + \sum_{j=1}^c b_j^2)\big ) \\
%k_{10} & =  \frac{1}{2} \big( \sum_{i=1}^r a_i^2 - \sum_{i=1}^r \sum_{j=1}^c n_{ij}^2 \big)\\
%k_{01} & = \frac{1}{2} \big ( \sum_{j=1}^c b_j^2- \sum_{i=1}^r \sum_{j=1}^c n_{ij}^2\big)
%\end{align*}
And $k_{11}+k_{10}+k_{01}+k_{00} = {N \choose 2}$.
The sum of $k_{11} + k_{00}$ is interpreted as the total number of \emph{agreements} between $U$ and $V$, and the sum $k_{10} + k_{01}$ is the total number of disagreements. 
External CVIs based on pair-counting are computed with these four types of pairs.  Please refer to Table~\ref{tb:RIformu} for a non-exhaustive list of $26$ popular pair-counting based external CVIs~\citep{jim2010,lfamily2006}. These are the indices which will be discussed in terms of their susceptibility to NC bias and GT bias.

\begin{center}
\renewcommand{\arraystretch}{0.82}
\begin{longtable}{|p{0.03\textwidth}|p{.32\textwidth} | p{.1\textwidth}| p{.5\textwidth}|p{.07\textwidth}|}
   \caption{Pair-Counting based Comparison Measures (external CVIs)} \\
	\hline 
 $\#$	& \textbf{Name/Reference } & \textbf{Symbol} & \textbf{Formula} & \textbf{Find} \\
	\hline
	\endfirsthead
	\multicolumn{5}{c}%
	{\textit{Continued from previous page}} \\
	\hline
 $\#$	& \textbf{Name/Reference } & \textbf{Symbol} & \textbf{Formula} & \textbf{Find} \\
	\hline
	\endhead
	\endfoot
    \hline
	\endlastfoot
	\hline
	$1$ &  Rand Index~\citep{rand1971} & RI & $\frac{k_{11}+k_{00}}{k_{11}+k_{10}+k_{01}+k_{00}}$ & Max \\
	\hline
	$2$ & Adjusted Rand Index& \multirow{2}{*}{ARI} & \multirow{2}{*}{$\frac{k_{11} - \frac{(k_{11}+k_{10})(k_{11}+k_{01})}{k_{11}+k_{10}+k_{01}+k_{00}}} {\frac{(k_{11}+k_{10})+(k_{11}+k_{01})}{2} - \frac{(k_{11}+k_{10})(k_{11}+k_{01})}{k_{11}+k_{10}+k_{01}+k_{00}}}$} & \multirow{2}{*}{Max} \\
	& Hubert and Arabie~\citep{hubert1985ari} & & & \\
	\hline
	$3$ & Mirkin ~\citep{mirkin1996} & Mirkin & $2(k_{10}+k_{01})$ & Min\\
	\hline 
	$4$ & Jaccard Index~\citep{jaccard1908} & JI & $\frac{k_{11}}{k_{11}+k_{10}+k_{01}}$& Max \\
	\hline
	$5$ & Hubert~\citep{hubert1977} & H & $\frac{(k_{11}+k_{00}) - (k_{10}+k_{01})}{k_{11}+k_{10}+k_{01}+k_{00}}$& Max \\
	\hline
	$6$ & Wallace~\citep{wallace1983} & W1 & $\frac{k_{11}}{k_{11}+k_{10}}$ & Max \\
	\hline
	$7$ & Wallace~\citep{wallace1983} & W2 & $\frac{k_{11}}{k_{11} + k_{01}}$ & Max\\ 
	\hline 
	$8$ & Fowlkes \&  Mallow~\citep{fowlkes1983} & FM & $\frac{k_{11}}{\sqrt{(k_{11}+k_{10})(k_{11}+k_{01})}}$& Max \\
	\hline
	$9$ & Minkowski~\citep{minkowski2004} & MK & $\sqrt{\frac{k_{10}+k_{01}}{k_{11}+k_{10}}}$ & Min\\ 
	\hline 
	$10$ & Hubert's Gamma~\citep{jain1988} & $\Gamma$ & $\frac{k_{11}k_{00} - k_{10}k_{01}}{\sqrt{(k_{11}+k_{10})(k_{11}+k_{01})(k_{01}+k_{00}) (k_{10}+k_{00})}}$ & Max\\
	\hline
	$11$ & Yule~\citep{sneath1973} & Y & $\frac{k_{11}k_{00} - k_{10}k_{01}}{k_{11}k_{10} + k_{01}k_{00}}$ & Max\\ 
	\hline
	$12$ & Dice~\citep{dice1945} & Dice & $\frac{2k_{11}}{2k_{11}+k_{10}+k_{01}}$ & Max\\ 
	%& Dice~\citep{dice1945} & & & \\ 
%	$12$ & Czekanowski~\citep{czekanowski1932} & \multirow{2}{*}{Dice} & \multirow{3}{*}{$\frac{2k_{11}}{2k_{11}+k_{10}+k_{01}}$} & \multirow{3}{*}{Max}\\ 
%	& Dice~\citep{dice1945} & & & \\ 
	%& Gower \& Legendre~\citep{gower1986} & & &\\
	\hline 
	$13$ & Kulczynski~\citep{kulczynski1928} & K & $\frac{1}{2} \big( \frac{k_{11}}{k_{11}+k_{10}} + \frac{k_{11}}{k_{11}+k_{01}} \big)$ & Max\\
	\hline
	$14$ & McConnaughey~\citep{mcconnaughey1964} & MC &$\frac{k_{11}^2 - k_{10}k_{01}}{(k_{11}+k_{10}) (k_{11}+k_{01})}$ & Max\\
	\hline
	$15$ & Peirce~\citep{peirce1884} & PE & $\frac{k_{11}k_{00} - k_{10}k_{01}}{(k_{11}+k_{01})(k_{10}+k_{00})}$ & Max \\
	\hline
	$16$ & Sokal \& Sneath~\citep{sokal1963} & SS1 & $\frac{1}{4} \big(\frac{k_{11}}{k_{11}+k_{10}} + \frac{k_{11}}{k_{11}+k_{01}} +\frac{k_{00}}{k_{10}+k_{00}} + \frac{k_{00}}{k_{01} + k_{00}}\big)$ & Max\\
	\hline
	$17$ &　Baulieu~\citep{baulieu1989} & B1 & $\frac{{N \choose 2}^2 - {N \choose 2} (k_{10} + k_{01}) + (k_{10} - k_{01})^2}{{N \choose 2}^2}$ & Max\\
	\hline
	$18$ & Russel \& Rao~\citep{russell1940} & RR & $\frac{k_{11}}{k_{11}+k_{10}+k_{01}+k_{00}}$ & Max\\
	\hline
	$19$ & Fager \& McGowan~\citep{fager1963} & FMG & $\frac{k_{11}}{\sqrt{(k_{11}+k_{10}) (k_{11}+k_{01})}} - \frac{1}{2\sqrt{k_{11}+k_{10}}}$  & Max\\
	\hline
	$20$ & Pearson & P & $\frac{k_{11}k_{00} - k_{10}k_{01}}{(k_{11}+k_{10}) (k_{11}+k_{01}) (k_{01} + k_{00}) (k_{10}+k_{00})}$  & Max\\
	\hline
	$21$ & Baulieu~\citep{baulieu1989} & B2 & $\frac{k_{11}k_{00} - k_{10}k_{01}}{{N \choose 2
	}^2}$ & Max\\
	\hline
	$22$ & Sokal \& Sneath~\citep{sokal1963} & SS2 & $\frac{k_{11}}{k_{11} + 2(k_{10} + k_{01})}$ & Max\\
	\hline
	$23$ & Sokal \& Sneath~\citep{sokal1963} & \multirow{2}{*}{SS3} & \multirow{2}{*}{$\frac{k_{11}k_{00}}{\sqrt{(k_{11}+k_{10}) (k_{11}+k_{01}) (k_{10} + k_{00}) (k_{01}+k_{00})}}$} & \multirow{2}{*}{Max}\\ 
	& Ochiai~\citep{ochiai1957} & & & \\
	\hline
	$24$ & Gower \& Legendre~\citep{gower1986} & \multirow{2}{*}{GL} & \multirow{2}{*}{$\frac{k_{11}+k_{00}}{k_{11} + \frac{1}{2}(k_{10} + k_{01}) + k_{00}}$} & \multirow{2}{*}{Max} \\ 
	& Sokal \& Sneath~\citep{sokal1963} & & &\\
	\hline
	$25$ & Rogers \& Tanimoto~\citep{rogers1960} & RT & $\frac{k_{11} + k_{00}}{k_{11} + 2(k_{10} + k_{01}) +k_{00}}$ & Max\\
	\hline
	$26$ &  \small{Goodman} \& \small{Kruskal}~\citep{goodman1954} & \multirow{2}{*}{GK} & \multirow{2}{*}{$\frac{k_{11}k_{00} - k_{10}k_{01}}{k_{11}k_{00} + k_{10}k_{01}}$} & \multirow{2}{*}{Max}\\ 
	& Yule~\citep{yule1900} & & & 
	%\\ 
	%\hline
\label{tb:RIformu}
\end{longtable} 
\end{center}

\section{Numerical Experiments} \label{sec:test}
In this section, we test and discuss $26$ pair-counting based external cluster validity indices listed in Table~\ref{tb:RIformu} with respect to NC bias, GT1 bias and GT2 bias. And we found that RI and $4$ related CVIs show GT1 and GT2 bias behaviour. 
\subsection{Type 1: GT1 bias Testing} \label{subsec:gt1t}
\begin{figure}[ht!]
\centering
\setlength{\fboxsep}{1pt}
\setlength{\fboxrule}{1pt}
\framebox[1\width][c]{
\subfloat[RI with random $U_{GT} \in M_{h2N}$, i.e., $c_{true}=2$.]{
\begin{tikzpicture}[squarednode/.style={rectangle, draw=red!60, fill=red!5, very thick, minimum size=5mm},]
\draw(0,0) node[inner sep=0]{\includegraphics[width=0.4\textwidth]{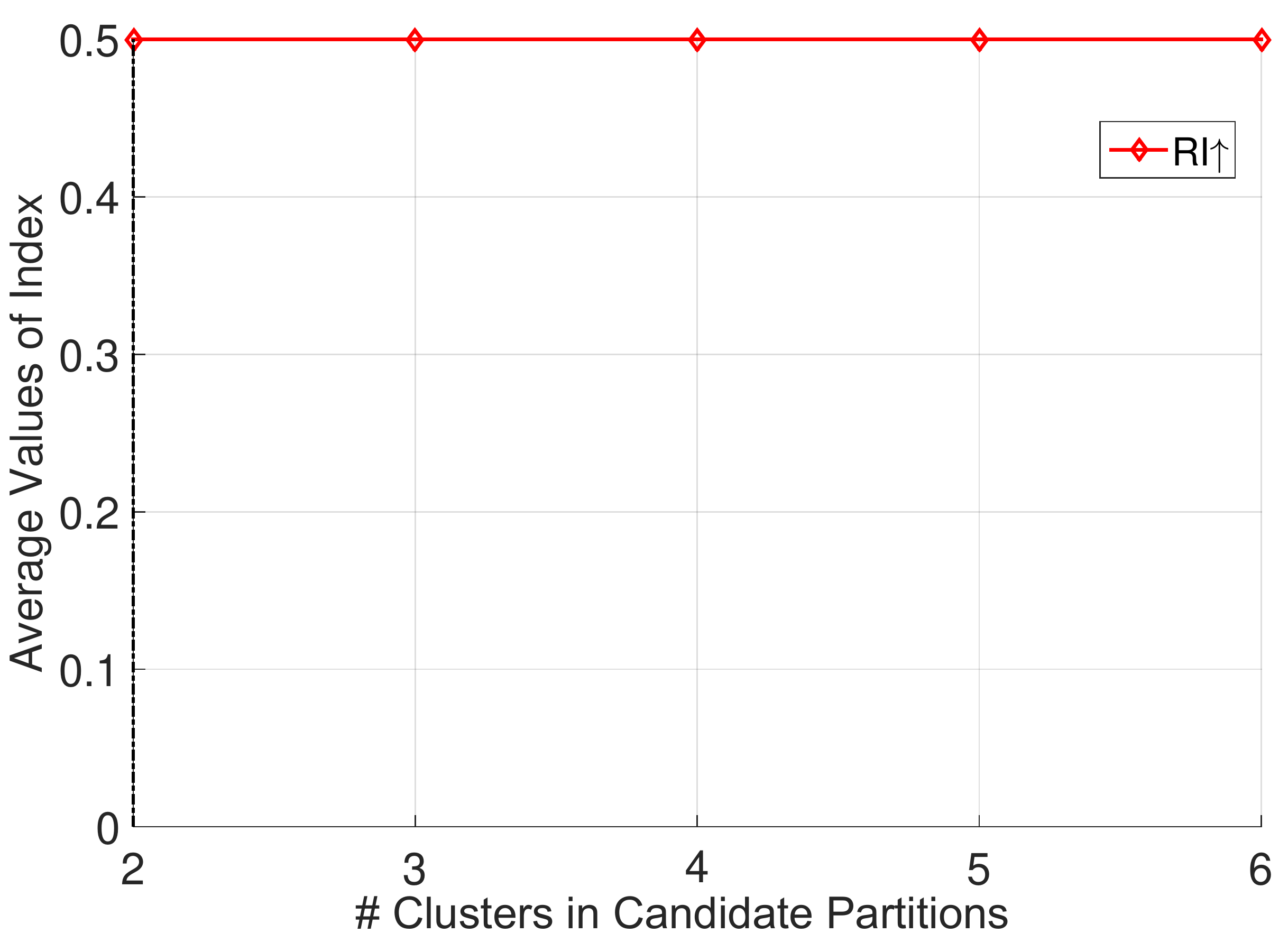}
\label{subfig:ri2c}};
\draw(1,0) node[squarednode](s-RI2){NCneu};
\end{tikzpicture}
}
\quad
\subfloat[RI with random $U_{GT} \in M_{h50N}$, i.e., $c_{true}=50$.]{
\begin{tikzpicture}[squarednode/.style={rectangle, draw=red!60, fill=red!5, very thick, minimum size=5mm},]
\draw(0,0) node[inner sep=0] {\includegraphics[width=0.4\textwidth]{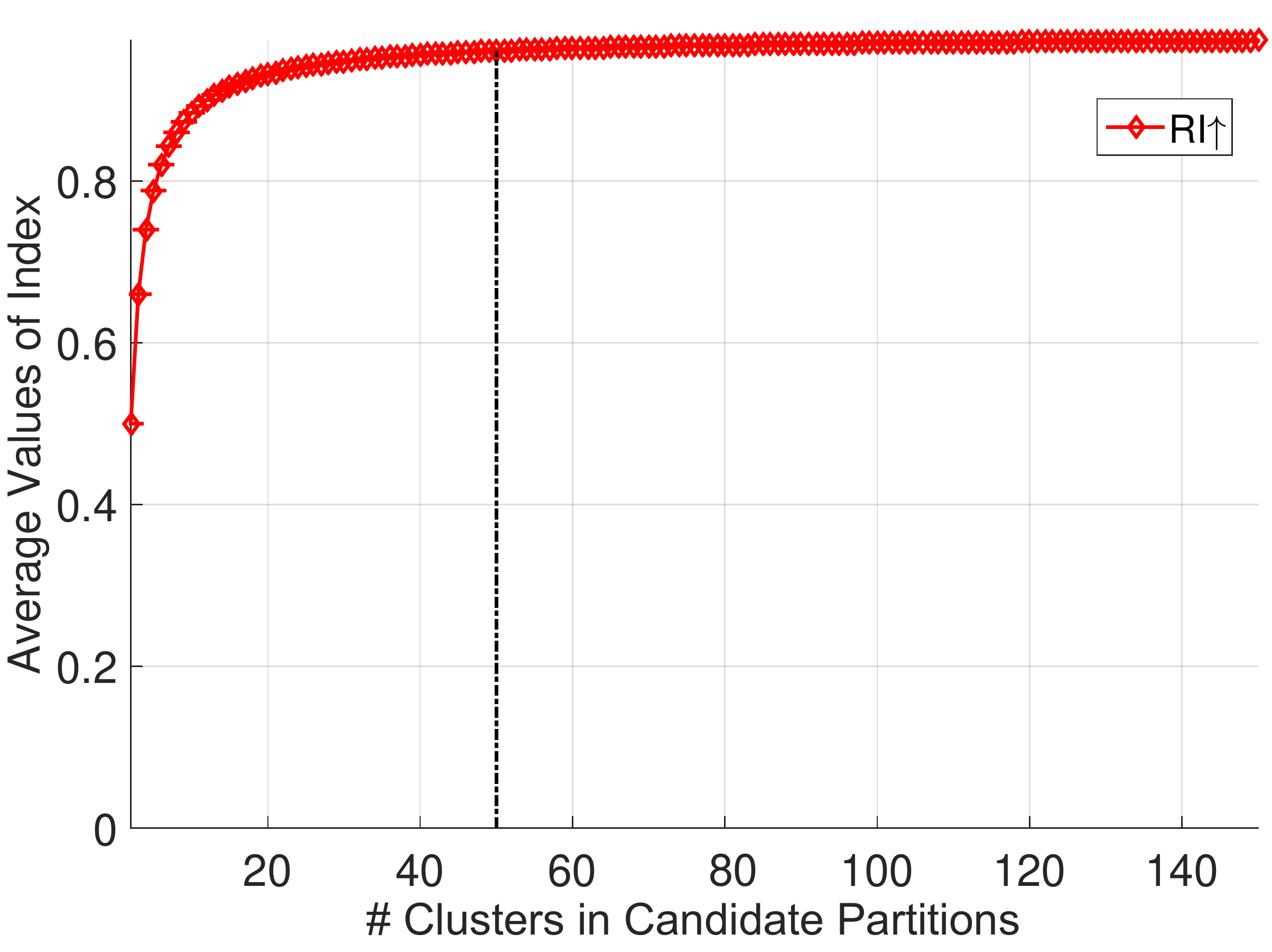}
\label{subfig:ric50}};
\draw(0,0) node[squarednode](d-RI50){NCinc};
\end{tikzpicture}
}
\tikz[remember picture, overlay] \draw[line width=1pt,-stealth,red] ([xshift=0mm]s-RI2.east) -- ([xshift=0mm]d-RI50.west)node[midway,above,text=black,font=\large\bfseries\sffamily] {$\Rightarrow$GT1 bias};
}

\vspace{0.2cm}
\framebox[1\width][c]{
\subfloat[Jaccard with random $U_{GT} \in M_{h2N}$, i.e., $c_{true}=2$.]{ 
\begin{tikzpicture}[squarednode/.style={rectangle, draw=red!60, fill=red!5, very thick, minimum size=5mm},]
\draw(0,0) node[inner sep=0]{\includegraphics[width=0.4\textwidth]{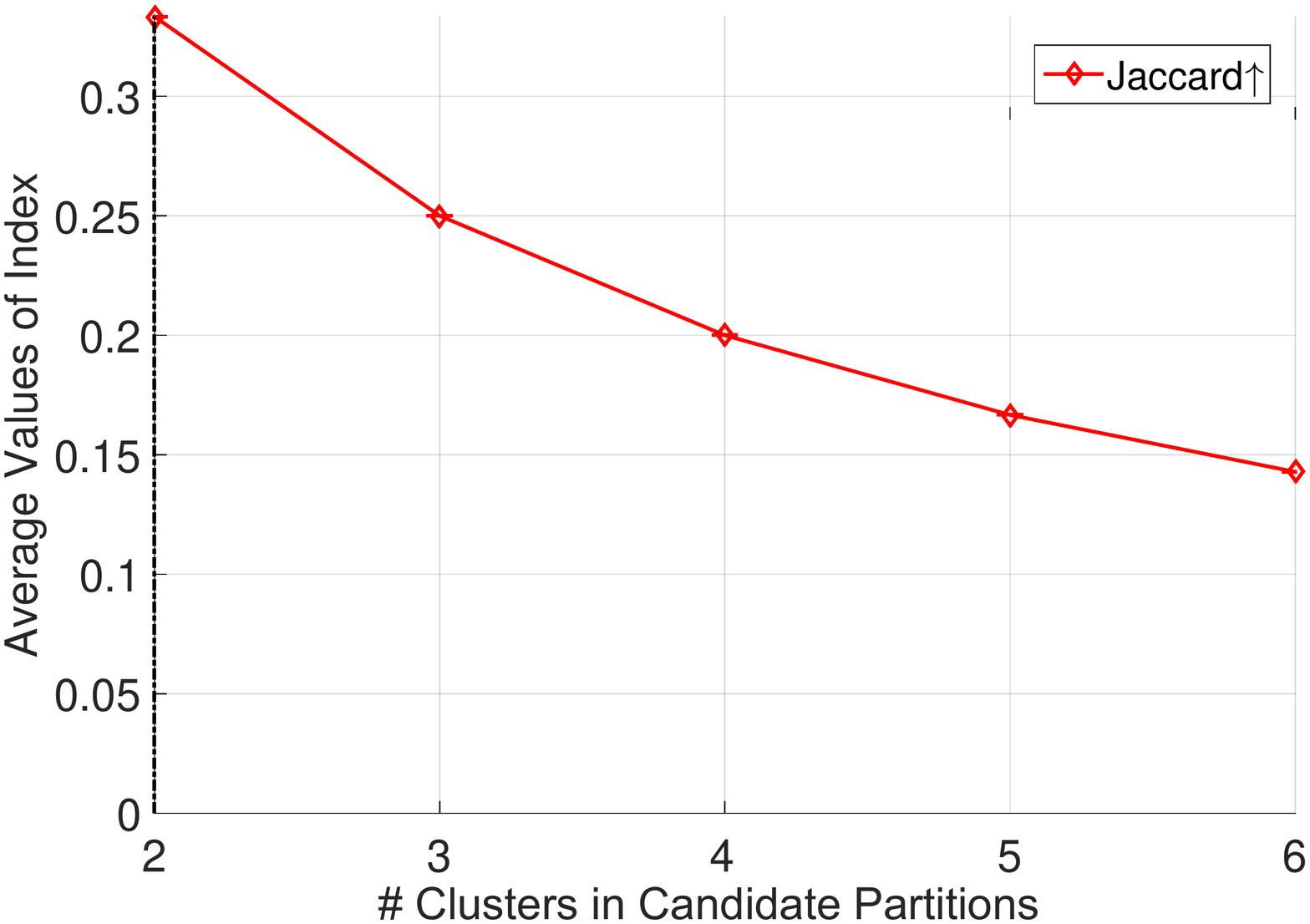}
\label{subfig:jc2}};
\draw(1,-0.5) node[squarednode]{NCdec};
\end{tikzpicture}
}
\quad
\subfloat[Jaccard with random $U_{GT} \in M_{h50N}$, i.e., $c_{true}=50$.]{
\begin{tikzpicture}[squarednode/.style={rectangle, draw=red!60, fill=red!5, very thick, minimum size=5mm},]
\draw(0,0) node[inner sep=0]{\includegraphics[width=0.4\textwidth]{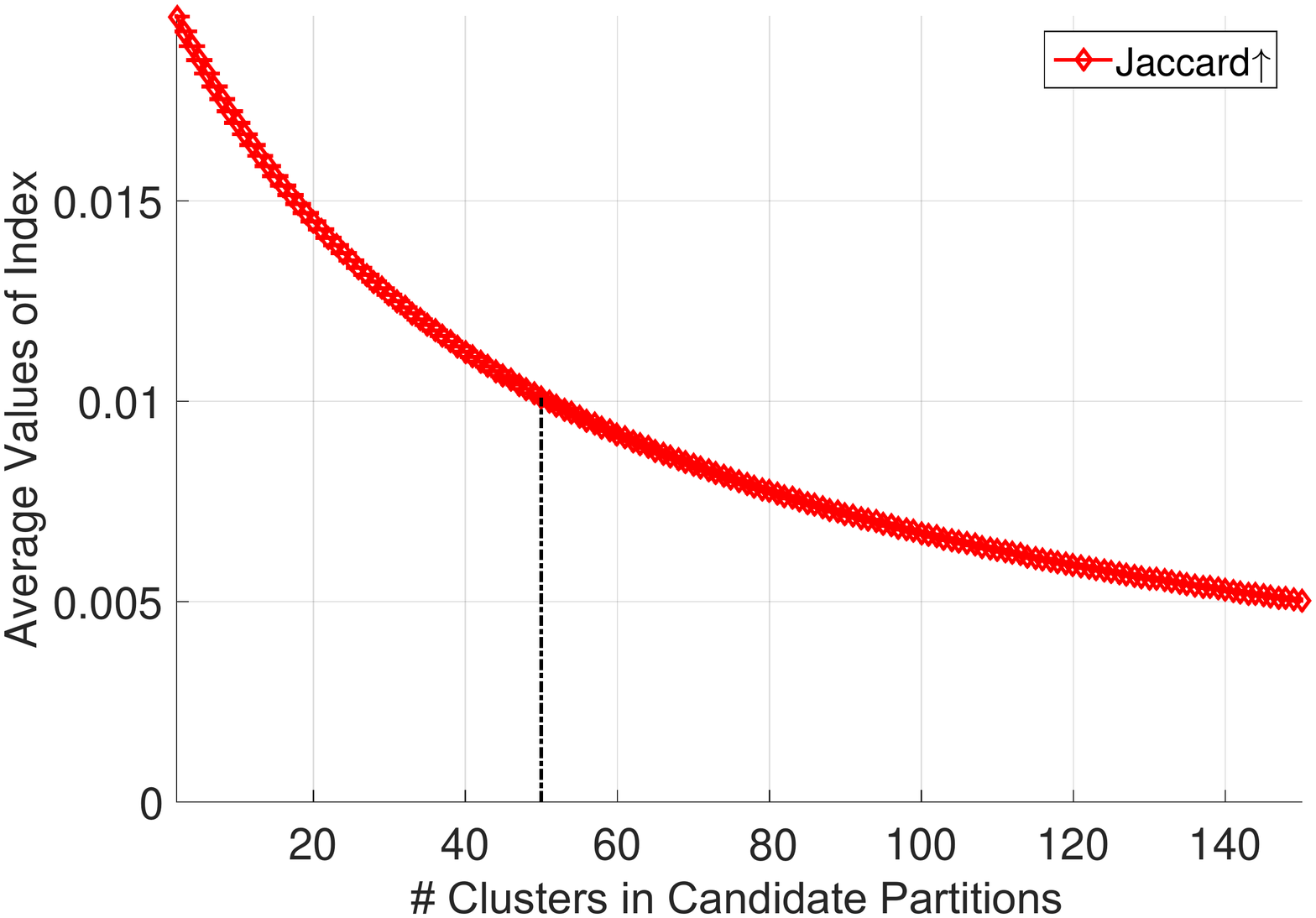}
\label{subfig:jc50}};
\draw(0.2,-0.6) node[squarednode]{NCdec};
\end{tikzpicture}
}
}

\vspace{0.2cm}
\framebox[1\width][c]{
\subfloat[ARI with random $U_{GT} \in M_{h2N}$, i.e., $c_{true}=2$.]{
\begin{tikzpicture}[squarednode/.style={rectangle, draw=red!60, fill=red!5, very thick, minimum size=5mm},]
\draw(0,0) node[inner sep=0]{\includegraphics[width=0.4\textwidth]{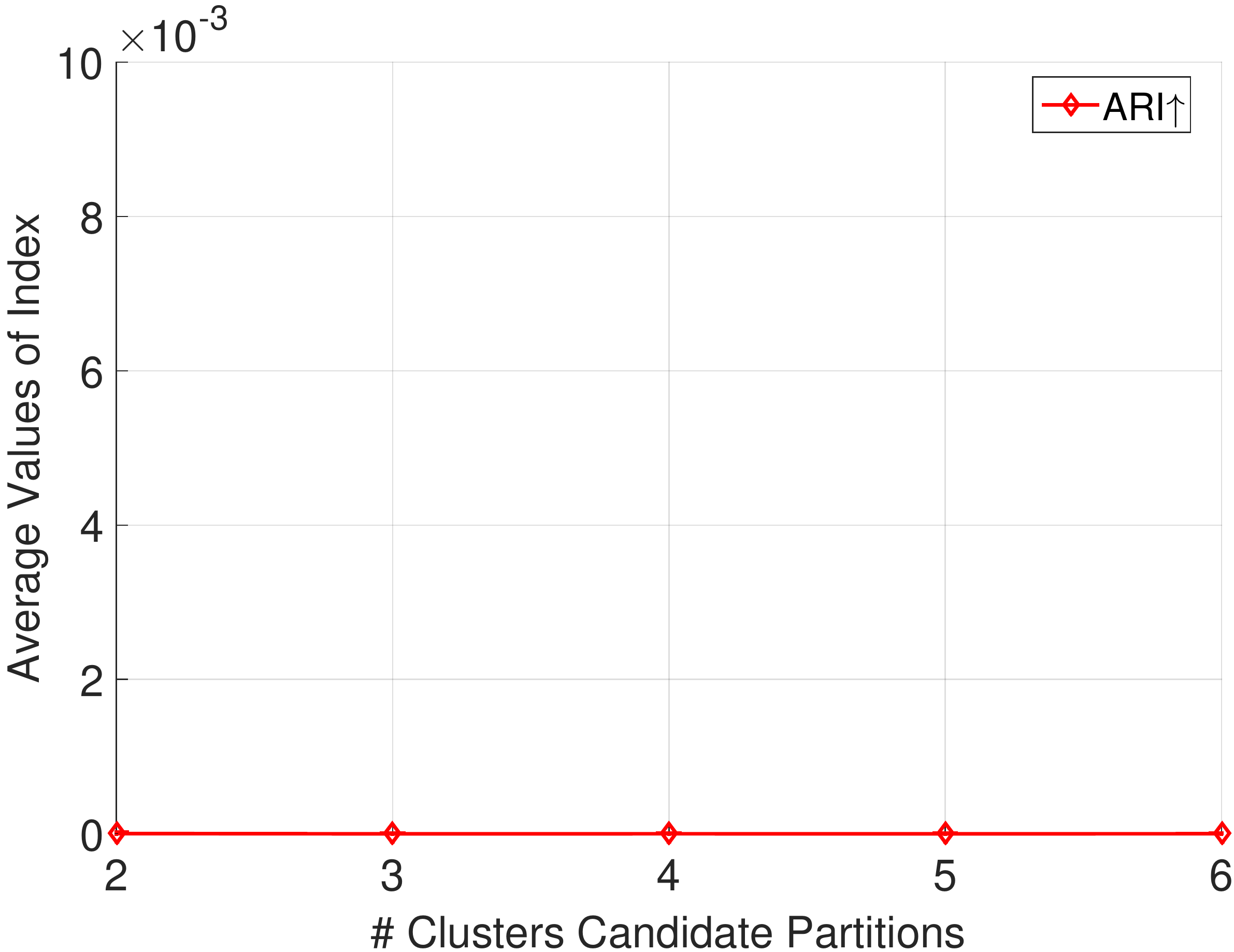}
\label{subfig:aric2}};
\draw(1,0) node[squarednode]{NCneu};
\end{tikzpicture}
}
\quad
\subfloat[ARI with random $U_{GT} \in M_{h50N}$, i.e., $c_{true}=50$.]{
\begin{tikzpicture}[squarednode/.style={rectangle, draw=red!60, fill=red!5, very thick, minimum size=5mm},]
\draw(0,0) node[inner sep=0]{\includegraphics[width=0.4\textwidth]{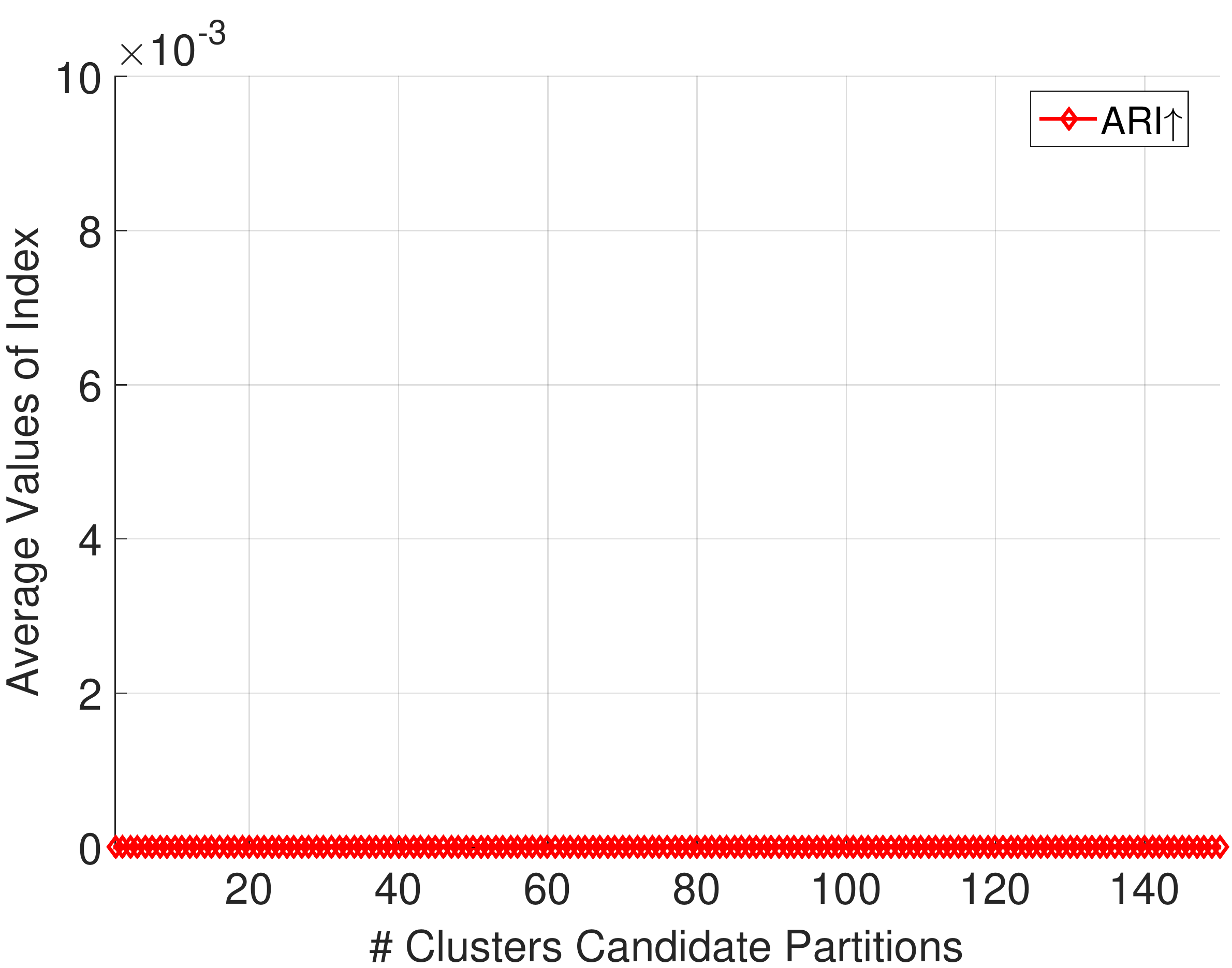}
\label{subfig:aric50}};
\draw(0.2,0) node[squarednode]{NCneu};
\end{tikzpicture}
}
}
\caption{$100$ trial average values of the RI, JI and ARI external CVIs with variable ground truth resulting in GT1 bias, $c_{true} = 2,50$.}
\label{fig:pcnum}
\end{figure}

We use the same experimental setting as in Example $1$. 
The ground truth partition $U_{GT}$ is randomly generated with $c_{true}$ subsets which are in each case uniformly distributed in size, where $c_{true} = \{2, 10, 20, 30, 50\}$. Then, we randomly generate $100$ candidate partitions with $c$ clusters, where $c$ ranges from $2$ to $3 * c_{true}$.  We performed this experiment on all $26$ comparison measures shown in Table~\ref{tb:RIformu}, but due to limited space, we focus our discussion on the results from three representative measures, the RI, JI and ARI (indices $\#1, \#2, and~ \#4$ in Table~\ref{tb:RIformu}) with $c_{true} = 2, 50$ (Figure~\ref{fig:pcnum}).

When $c_{true} = 2$, the RI trend is flat, that is, it has NCneu bias. But when $c_{true} = 50$, the RI favors solutions with larger number of clusters, i.e., it shows NCinc bias with $c_{true}=50$. Thus, the number of clusters in the random ground truth partition $U_{GT}$ does influence the NC bias behaviour of the RI. According to definition~\ref{def:gt1bias}, this indicates that RI has GT1 bias.
Comparing Figures~\ref{subfig:jc2} and~\ref{subfig:jc50} shows that the Jaccard index does not seem to suffer from GT bias due to the number of subsets in $U_{GT}$. These two figures show that the JI exhibits NCdec bias, decreasing monotonically as c increases from $2$ to $6$ (Figure~\ref{subfig:jc2}) or $2$ to $150$ (Figure~\ref{subfig:jc50}). Figures~\ref{subfig:aric2} and~\ref{subfig:aric50} show that the ARI is not monotonic for either value of $c$, and is not affected by the number of clusters in $U_{GT}$. Thus, ARI has NCneu bias. We remark that these observed bias behaviours of the tested external CVIs are based on these experimental settings.

\subsection{Type 2: GT2 bias Testing} \label{subsec:gt2t}
\begin{figure}[ht!]
\centering
\setlength{\fboxsep}{1pt}
\setlength{\fboxrule}{1pt}
\framebox[1\width][c]{
\subfloat[RI with skewed ground truth $c_{true}=5$: $n_1 = 10\%*N$.]{
\begin{tikzpicture}[squarednode/.style={rectangle, draw=red!60, fill=red!5, very thick, minimum size=5mm},]
\draw(0,0) node[inner sep=0]{\includegraphics[width=0.4\textwidth]{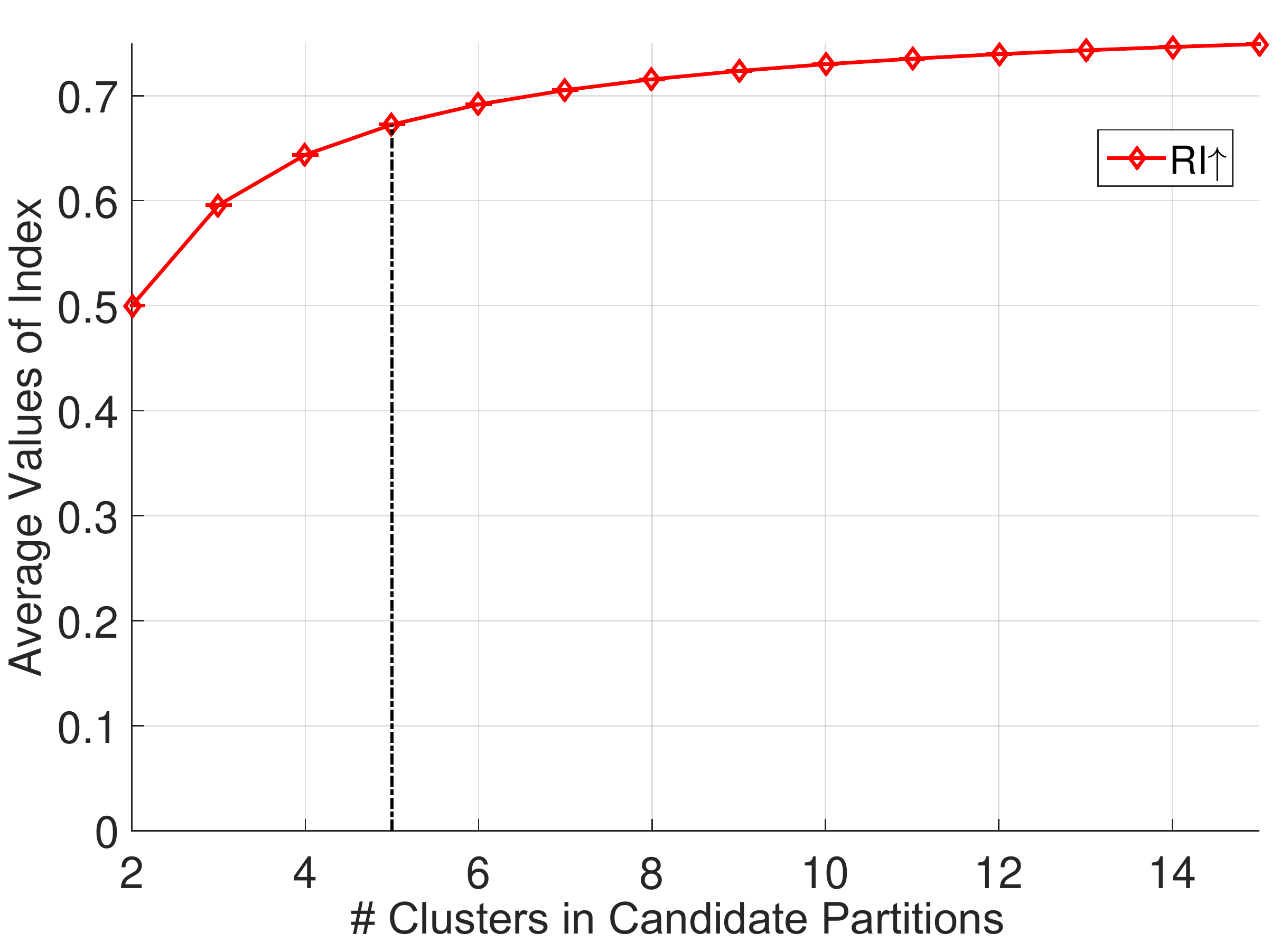}
\label{subfig:riskp1}};
\draw(1,-0.15) node[squarednode](s-RIp1){NCinc};
\end{tikzpicture}
}
\quad
\subfloat[RI with skewed ground truth $c_{true}=5$: $n_1 = 90\% * N$.]{
\begin{tikzpicture}[squarednode/.style={rectangle, draw=red!60, fill=red!5, very thick, minimum size=5mm},]
\draw(0,0) node[inner sep=0]{\includegraphics[width=0.4\textwidth]{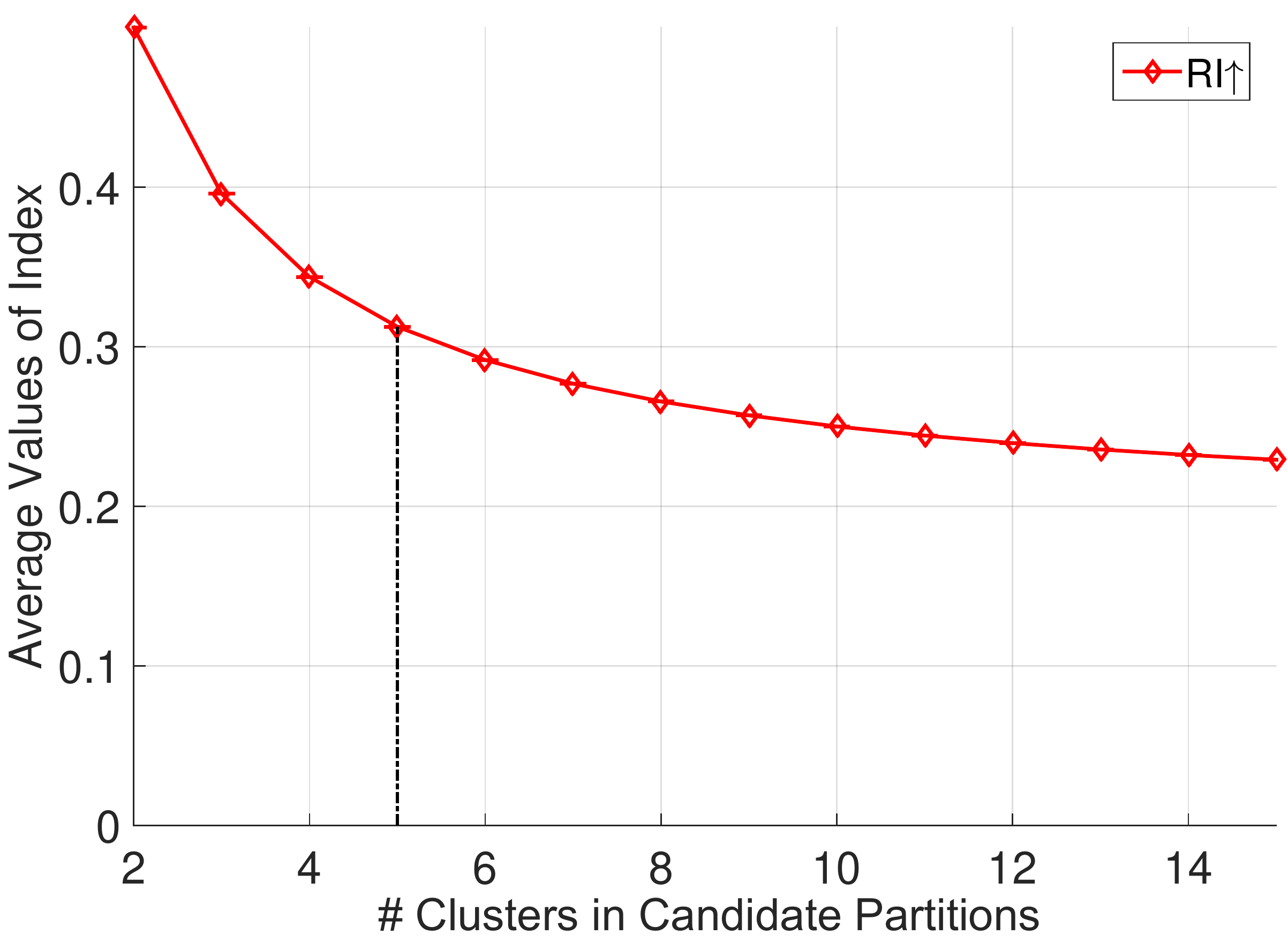}
\label{subfig:riskp8}};
\draw(0,-0.15) node[squarednode](d-RIp9){NCdec};
\end{tikzpicture}
}
\tikz[remember picture, overlay] \draw[line width=1pt,-stealth,red] ([xshift=0mm]s-RIp1.east) -- ([xshift=0mm]d-RIp9.west)node[midway,above,text=black,font=\large\bfseries\sffamily] {$\Rightarrow$GT2 bias};
}

\vspace{0.2cm}
\framebox[1\width][c]{
\subfloat[Jaccard with skewed ground truth $c_{true}=5$: $n_1 = 10\%*N$.]{
\begin{tikzpicture}[squarednode/.style={rectangle, draw=red!60, fill=red!5, very thick, minimum size=5mm},]
\draw(0,0) node[inner sep=0]{\includegraphics[width=0.4\textwidth]{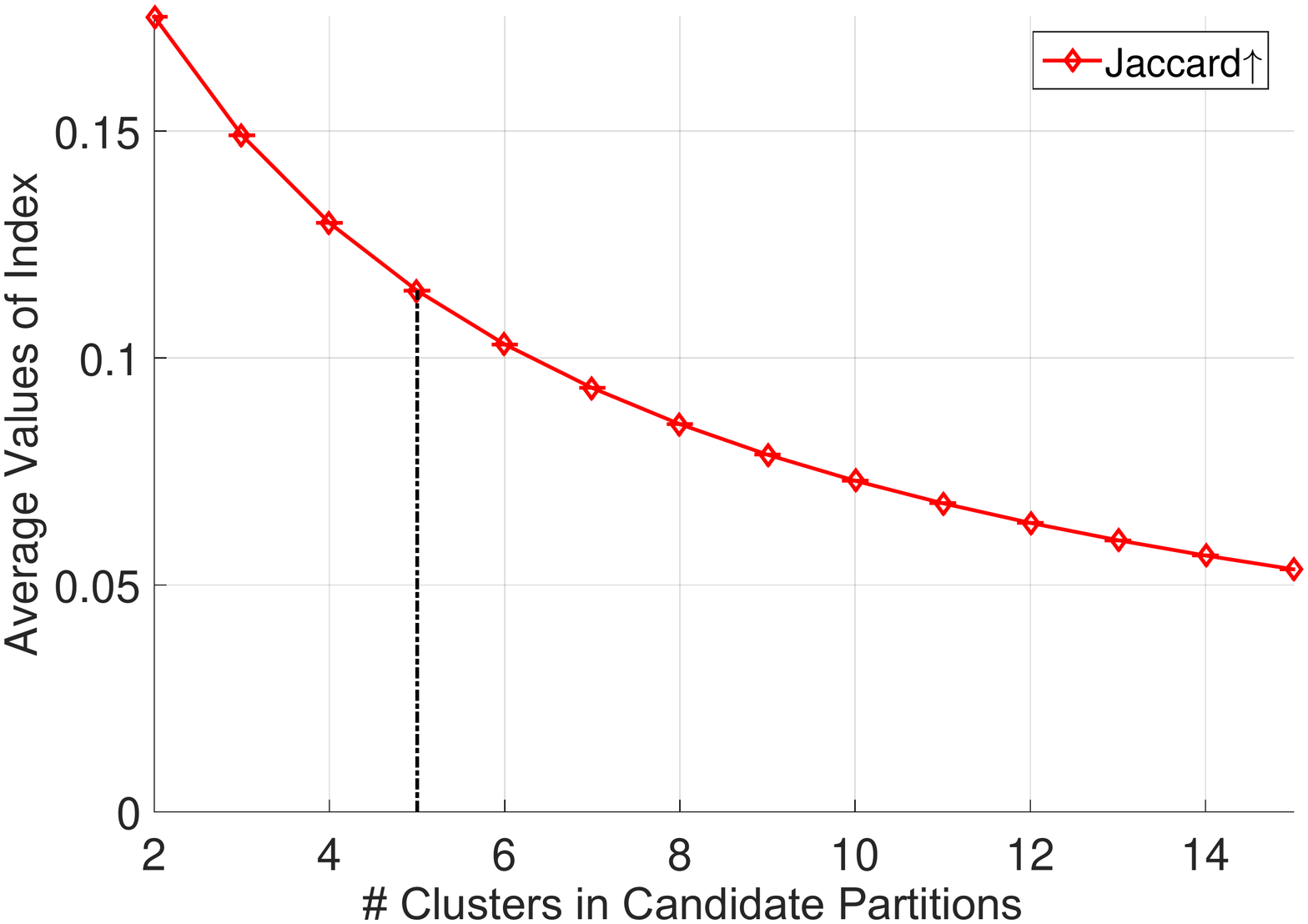}
\label{subfig:jskp1}};
\draw(1,0.5) node[squarednode]{NCdec};
\end{tikzpicture}
}
\quad
\subfloat[Jaccard with skewed ground truth $c_{true}=5$: $n_1 = 90\% * N$.]{
\begin{tikzpicture}[squarednode/.style={rectangle, draw=red!60, fill=red!5, very thick, minimum size=5mm},]
\draw(0,0) node[inner sep=0]{\includegraphics[width=0.4\textwidth]{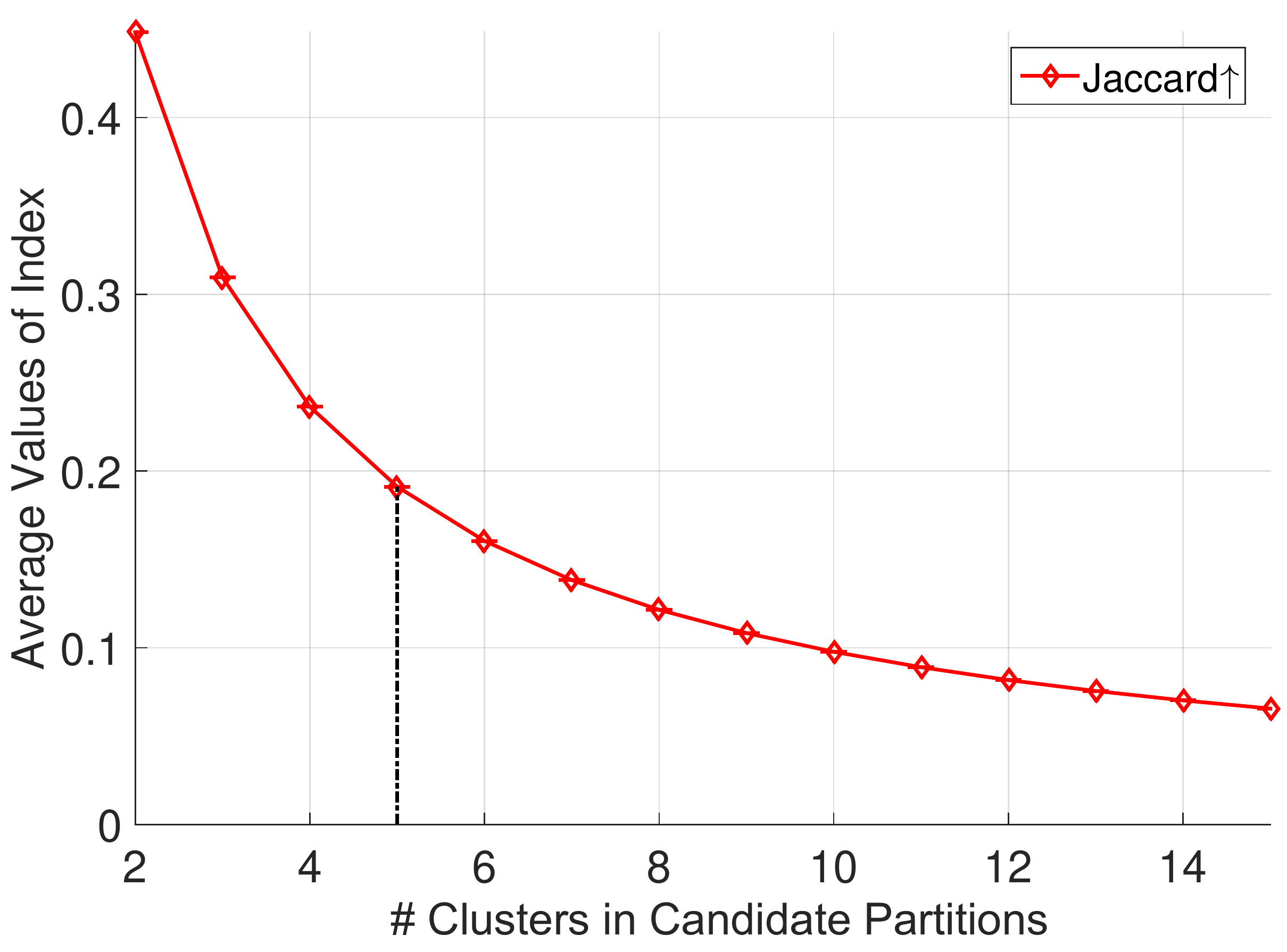}
\label{subfig:jskp8}};
\draw(0.2,0.5) node[squarednode]{NCdec};
\end{tikzpicture}
}
}

\vspace{0.2cm}
\framebox[1\width][c]{
\subfloat[ARI with skewed ground truth $c_{true}=5$: $n_1 = 10\%*N$.]{
\begin{tikzpicture}[squarednode/.style={rectangle, draw=red!60, fill=red!5, very thick, minimum size=5mm},]
\draw(0,0) node[inner sep=0]{\includegraphics[width=0.4\textwidth]{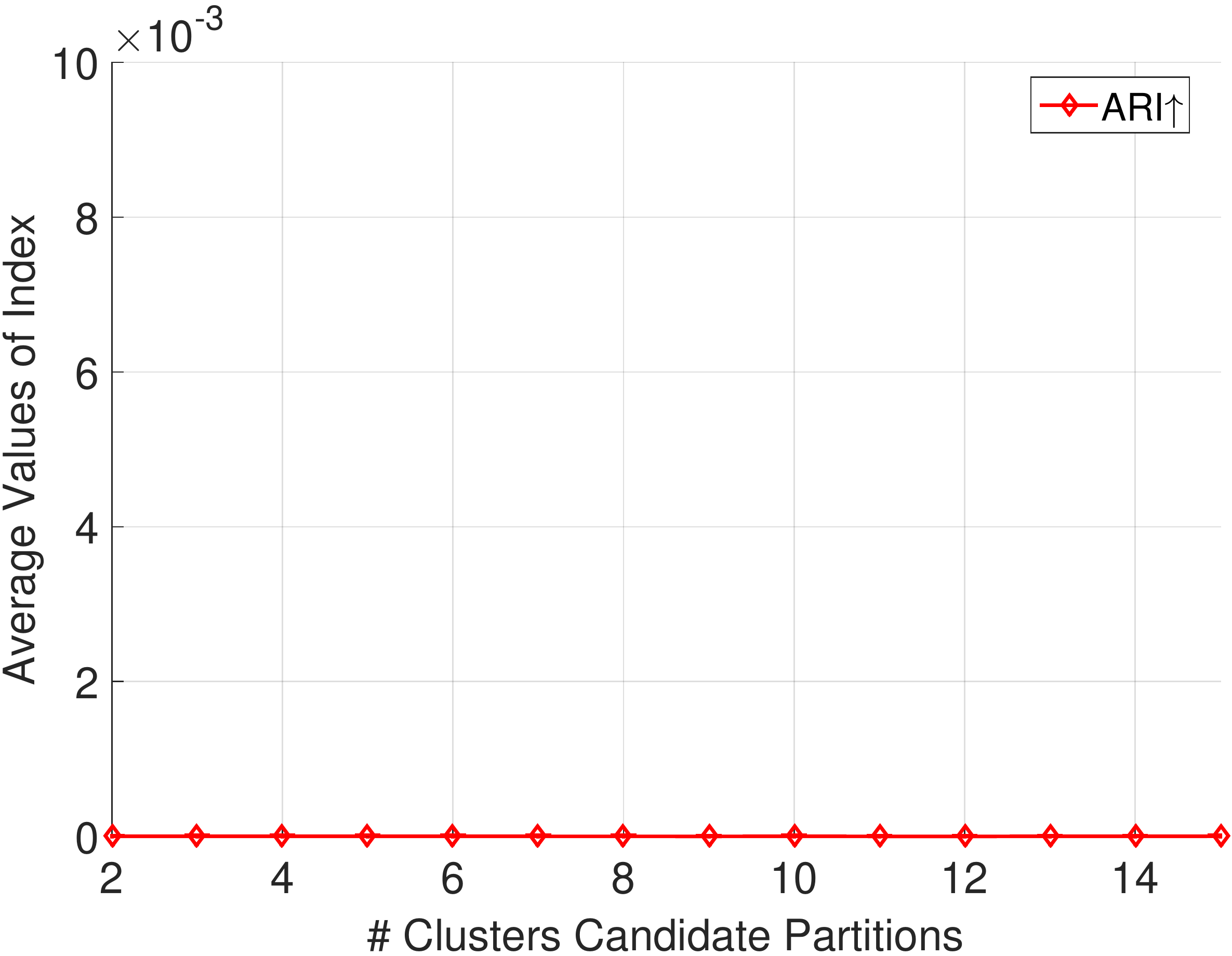}
\label{subfig:ariskp1}};
\draw(1,0) node[squarednode]{NCneu};
\end{tikzpicture}
}
\quad
\subfloat[ARI with skewed ground truth $c_{true}=5$: $n_1 = 90\% * N$.]{
\begin{tikzpicture}[squarednode/.style={rectangle, draw=red!60, fill=red!5, very thick, minimum size=5mm},]
\draw(0,0) node[inner sep=0]{\includegraphics[width=0.4\textwidth]{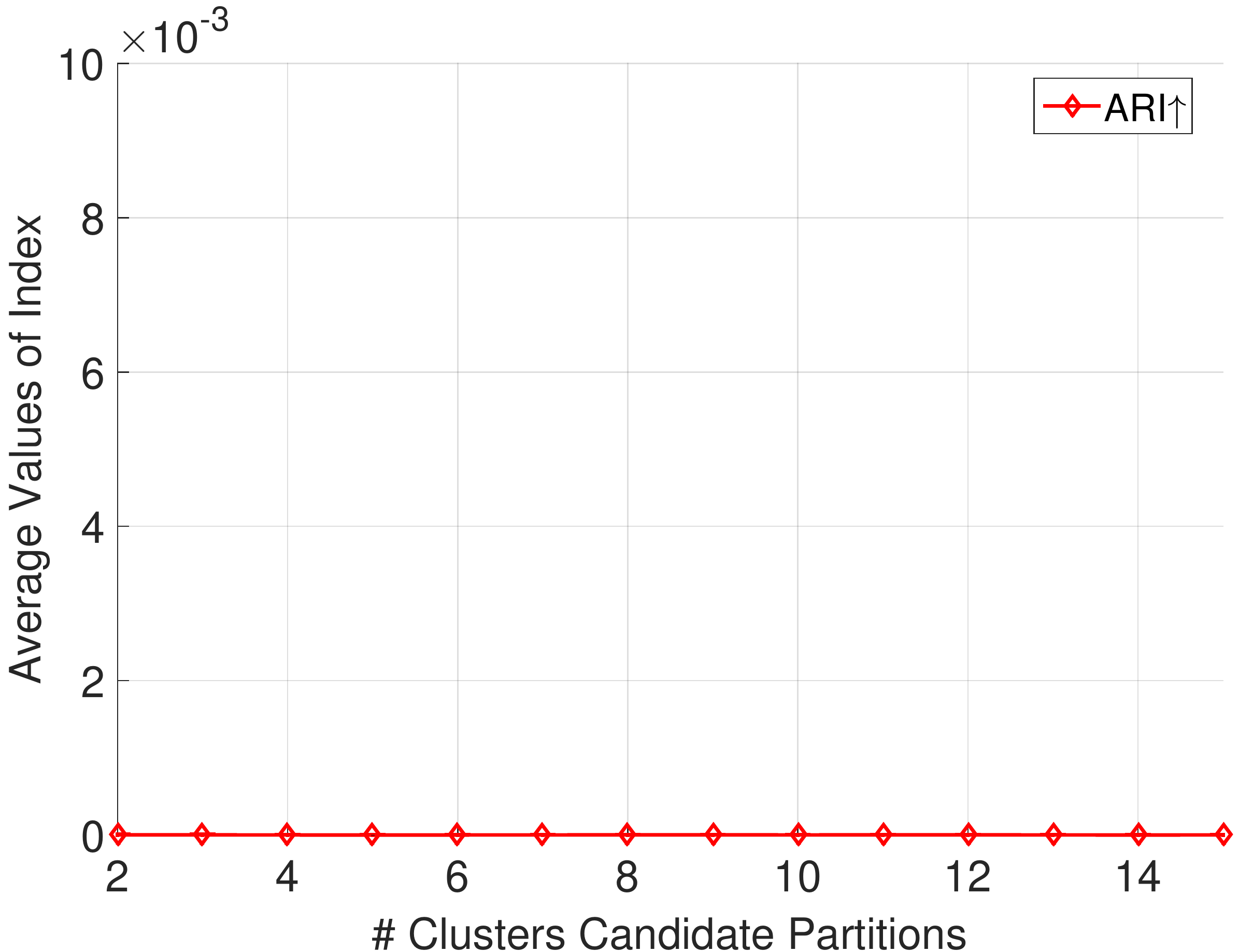}
\label{subfig:ariskp8}};
\draw(0.2,0) node[squarednode]{NCneu};
\end{tikzpicture}
}
}
\caption{$100$ run average values of the RI, JI and ARI with unbalanced ground truth to investigate GT2bias, $c_{true}=5$. $n_1 = 10\% * N$ (left), $n_1 = 90\% * N$ (right).}
\label{fig:pcskw}
\end{figure}

We use an experimental setup similar to that in Example $2$. We generate a ground truth by randomly assigning $10\%, 20\%, \ldots, 90\%$ of the objects to the first cluster, and then randomly assigning the remaining cluster labels to the rest of the data objects. Here $c_{true}=5$ is discussed.
%, and each of the remaining $4$ clusters contains an equal number of the remaining labels.
Figure~\ref{fig:pcskw} shows the results for the RI, JI and ARI with the size of the first cluster either $n_1 = 0.1*N$ or $n_1 = 0.9*N$. 

Figures~\ref{subfig:riskp1} and~\ref{subfig:riskp8} show that the RI suffers from GT2 bias according to definition~\ref{def:gt2bias}. It is monotone increasing with $n_1 = 10,000$ (NCinc bias), but monotone decreasing with $n_1 = 90,000$ (NCdec bias). Note that the graphs in Figures~\ref{subfig:riskp1} and~\ref{subfig:riskp8} are reflections of each other about the horizontal axis at $0.5$.
%The RI does not point to $c_{true}=5$ in either instance. 
The Jaccard index in Figures~\ref{subfig:jskp1} and~\ref{subfig:jskp8} exhibits the same NC bias status as it did in Figures~\ref{subfig:jc2} and~\ref{subfig:jc50}. Specifically, JI decreases monotonically with $c$, so it still has NCdec bias, but it does not seem to be affected by GT2 bias. 
%And again, it does not point to the preferred  value of $c$. 
The ARI in Figures~\ref{subfig:ariskp1} and~\ref{subfig:ariskp8} does not show any influence due to GT2 bias. It has NCneu bias under these two sets of experimental settings. So, from our empirical results, ARI would appear to be preferable to the RI and the JI in this setting. To summarize, these examples illustrate that the RI can suffer from GT1 bias and GT2 bias; that JI can suffer from NCdec bias but not GT1 bias nor GT2 bias; and that ARI does not suffer from NC bias or GT bias, under the experimental setup we have used here. 

\subsection{Summary for All 26 Comparison Measures}\label{subsec:5inds}
The overall results of similar experiments for all $26$ indices in Table~\ref{tb:RIformu} led to the conclusion that $5$ of the $26$ external CVIs suffer from GT1 bias and GT2 bias for these experimental settings. These measures are \emph{Rand Index} (RI) and

\begin{align}
%\text{Rand Index~\citep{rand1971}}:~RI(U,V) &=  RI \label{eq:ri} \\
\text{Hubert~\citep{hubert1977}}:~ H(U,V) & = 2RI - 1  \label{eq:Hri} \\
\text{Gower and Legendre~\citep{gower1986}}: ~ GL(U,V) &= \frac{2}{1+ 1/RI} \label{eq:GLri}\\
\text{Rogers and Tanimoto~\citep{rogers1960}}: ~RT(U,V) &= \frac{1}{2/RI - 1} \label{eq:RTri}\\
\text{Mirkin~\citep{mirkin1996}}: ~ Mirkin(U,V) &= N(N-1)(1-RI)\label{eq:Mirkin}
\end{align}

%\begin{align}
%\text{Rand Index~\citep{rand1971}}:~RI(U,V) & = \frac{k_{11}+k_{00}}{k_{11}+k_{00}+k_{10}+k_{01}} = RI \label{eq:ri} \\
%\text{Hubert~\citep{hubert1977}}:~ H(U,V) & = \frac{(k_{11}+k_{00}) - (k_{01}+k_{10})}{k_{11}+k_{00}+k_{10}+k_{01}}  = 2RI - 1  \label{eq:Hri} \\
%\text{Gower and Legendre~\citep{gower1986}}: ~ GL(U,V) &= \frac{k_{11}+k_{00}}{k_{11}+1/2(k_{10}+k_{01})+k_{00}}  = \frac{2}{1+\frac{1}{RI}} \label{eq:GLri}\\
%\text{Rogers and Tanimoto~\citep{rogers1960}}: ~RT(U,V) &= \frac{k_{11}+k_{00}}{k_{11} + 2(k_{10}+k_{01})+k_{00}} = \frac{1}{\frac{2}{RI} - 1} \label{eq:RTri}\\
%\text{Mirkin~\citep{mirkin1996}}: ~ M(U,V) &= 2(k_{10}+k_{01}) = N(N-1)(1-RI)\label{eq:Mirkin}
%\end{align}

Please note that the external CVIs in equations~\ref{eq:Hri},~\ref{eq:GLri},~\ref{eq:RTri} and~\ref{eq:Mirkin} are all functions of the RI. This observation forms the basis for our analysis in the next section.

\section{Bias Due to Ground Truth for the Rand Index} \label{sec:anly}
In this section, we provide a theoretical analysis for the GT bias, GT1 bias and GT2 bias for the Rand Index. More specifically, we will analyze the underlying reason for the GT bias of RI, based on its relationship with the quadratic entropy. Then, based on the relationship between RI and the quadratic entropy, we will discuss theoretically about when RI shows GT bias, GT1 bias and GT2 bias, according to the distribution of the ground truth and the number of subsets in the ground truth. 

\subsection{Quadratic Entropy and Rand Index}
The Havrda-Charvat entropy~\citep{havrda1967} is a generalization of the Shannon entropy. The quadratic entropy is the Havrda-Charvat generalized entropy with $\beta = 2$. 
\subsubsection{Havrda-Charvat Generalized Entropy}
The Havrda-Charvat generalized entropy for a crisp partition $U$ with $r$ clusters $U=\{u_1, \ldots,u_r\}$ is
\begin{equation}
 H_\beta(U) = \frac{1}{1- 2^{1-\beta}} (1 - \sum_{i=1}^r (\frac{|u_i|}{N})^\beta) \label{eq:hbeta0}
\end{equation}
where $\beta$ is any real number $>0$ and $\beta \neq 1$. Since $H$ is a continuous function of $\beta$, when $\beta=1$
\begin{equation}
 H_{1}(U) = - \sum_{i=1}^r \frac{|u_i|}{N} \log \frac{|u_i|}{N}
\end{equation}
which is the Shannon entropy $H_S(U)$.
When $\beta = 2$ we have quadratic entropy
\begin{equation}
H_2(U) = 2(1-\sum_{i=1}^r (\frac{|u_i|}{N})^2) \label{eq:h2u}
\end{equation}
It can be shown that in the case of statistically independent random variables $U$ and $V$
\begin{equation}
H_\beta(U,V) = H_\beta(U) + H_\beta(V) - (1-2^{1-\beta}) H_\beta(U) H_\beta(V) \label{eq:hbeta}
\end{equation}
When $\beta =2$, Equation~\ref{eq:hbeta} becomes 
\begin{equation}
H_2(U,V) = H_2(U) + H_2(V) - \frac{1}{2}H_2(U)H_2(V) \label{eq:h2uv}
\end{equation}

In~\citep{vi2007}, Meila showed that the \emph{Variation of Information} (VI) is a metric by expressing it as a function of Shannon's entropy. Consider a crisp partition $V$ with $c$ subsets $V=\{v_1,\ldots,v_c\}$, then
\begin{align}
VI (U,V)  &= H_S(U|V) + H_S(V|U) \label{eq:vishn}\\
          &= 2H_S(U,V) - H_S(U) - H_S(V) \nonumber
\end{align}
The VI is not one of the $26$ indices in Table~\ref{tb:RIformu}, but this information-theoretic CVI can be computed based on the contingency table, and it will help us analyze the GT bias of the $5$ external CVIs discussed in Section~\ref{subsec:5inds}.
%functions in Equations~\ref{eq:ri}-~\ref{eq:Mirkin}.

Simovici~\citep{generalized2007} showed that replacing Shannon's entropy in Equation~\ref{eq:vishn} by the generalized entropy at Equation~\ref{eq:hbeta0} still yielded a metric, 
\begin{align} 
VI_\beta(U,V) &= H_\beta(U|V) + H_\beta(V|U)  \\
                       &= 2H_\beta(U,V) - H_\beta(U) - H_\beta(V)\label{eq:vibeta}
\end{align}
For $\beta=2$, this becomes
\begin{equation}
VI_2(U,V) = 2H_2(U,V) - H_2(U) - H_2(V) \label{eq:vi2}
\end{equation}

Based on the above introduced concepts, we next introduce how to derive the relationship between RI and the quadratic entropy (i.e., Havrda-Charvat generalized entropy with $\beta=2$). This relationship will help us explain why RI shows GT bias.
\subsubsection{Quadratic Entropy vs. Rand Index}
Let $U$ and $V$ be two crisp partitions of $N$ samples with $r$ clusters and $c$ clusters respectively. Then the relationship between $VI_2(U,V)$ and $RI(U,V)$ can be derived as follows~\citep{generalized2007}.

First, based on Equations~\ref{eq:vibeta} and~\ref{eq:hbeta0}, we have $VI_\beta(U,V)$ as
\begin{align}
VI_\beta(U,V) &= 2H_\beta(U,V) - H_\beta(U) - H_\beta(V) \\
                       &= \frac{2}{1-2^{1-\beta}} (1 - \sum_{i=1}^r \sum_{j=1}^c (\frac{|u_i \cap v_j|}{N})^\beta) - \frac{1}{1-2^{1-\beta}}(1 - \sum_{i=1}^r\frac{|u_i|}{N})^\beta - \frac{1}{1-2^{1-\beta}} (1 - \sum_{j=1}^c(\frac{|v_j|}{N})^\beta) \nonumber \\
                       &= \frac{1}{1-2^{1-\beta}}(2 (1 - \sum_{i=1}^r \sum_{j=1}^c (\frac{|u_i \cap v_j|}{N})^\beta) - (1-\sum_{i=1}^r (\frac{|u_i|}{N})^\beta) - (1 - \sum_{j=1}^c (\frac{|v_j|}{N})^\beta)) \nonumber \\
                       &= \frac{1}{N^\beta(1-2^{1-\beta})} (\sum_{i=1}^r (|u_i|)^\beta + \sum_{j=1}^c (|v_j|)^\beta - 2 \sum_{i=1}^r \sum_{j=1}^c (|u_i \cap v_j|)^\beta)    \nonumber        
\end{align}
%\end{center}
Now setting $\beta = 2$, we get
\begin{center}
\begin{align}
VI_2(U,V) &= \frac{2}{N^2} (\sum_{i=1}^r (|u_i|)^2 + \sum_{j=1}^c (|v_j|)^2 - 2\sum_{i=1}^r \sum_{j=1}^c (|u_i \cap v_j|)^2) \nonumber \\
                 &= \frac{2}{N^2} (2k_{10} + 2k_{01}) \nonumber \\
                 &= \frac{2}{N} (N-1)(1-\mathbf{RI}(U,V)) \label{eq:vi2ri}
\end{align}
\end{center}
% \end{proof}
Equation~\ref{eq:vi2ri} shows that $VI_2$ and RI are inversely related. Thus, by analyzing the bias behaviour of $VI_2$,  it will be easy to understand the behaviour of RI. Next, we will analyze the GT bias behaviour of $VI_2$ based on the concept of quadratic entropy.

\subsection{GT bias of RI}
In this section, we will first discuss the general case of GT bias for RI by providing a series of theoretical statements for helping understand why RI shows GT bias, and when RI shows GT bias. Then, we will discuss two specific cases, i.e., GT1 bias and GT2 bias for RI and provide related theoretical statements which will explain when RI shows GT1 bias and GT2 bias. The related proofs are provided in~\ref{apdx}.
\subsubsection{General Case of GT bias}
We introduce Lemma~\ref{lema:vi2ri} to build the foundation for analyzing the GT bias of $VI_2$, then $RI$.
\begin{lemma} \label{lema:vi2ri}
Given $U \in M_{hrN}$ and $V \in M_{hcN}$, two statistically independent crisp partitions of $N$ data objects,  we have
\begin{equation}
VI_2(U,V) = H_2(U) + (1 - H_2(U))H_2(V) \label{eq:vih2}
\end{equation} 
\end{lemma}

Next, we introduce an important theorem in this paper that demonstrates why RI shows GT bias and when it shows GT bias by judging the relationship between the quadratic entropy of ground truth $U_{GT}$, $H_2(U_{GT})$ and $1$.
\begin{theorem} \label{thm:rih2u}
Let $U_{GT} \in M_{hrN}$ be a ground truth partition with $r$ subsets, and let $CP = \{V_1, \ldots, V_m\}$ be a set of candidate partitions with different numbers of clusters, 
where $ V_i \in M_{hc_iN}$ contains $c_i$ clusters which are uniformly distributed (balanced), $2 \leq c_i \leq N$.   Assuming $U_{GT}$ and $V_i \in CP$ are statistically independent, then RI suffers from GT bias. In addition, according to the relationship between $H_2(U_{GT})$ and $1$, we have: 
\begin{enumerate}
\item if $H_2(U_{GT})<1$, RI suffers from NCdec bias (i.e., RI decreases as $c_i$ increases);
\item If $H_2(U_{GT})=1$, RI is unbiased, i.e., NCneu bias (i.e., RI has no preferences as $c_i$ increases);
\item if $H_2(U_{GT})>1$, RI suffers from NCinc bias (i.e., RI increases as $c_i$ increases).
\end{enumerate}
\end{theorem}

\begin{figure}[t!]
\centering
\includegraphics[width=.8\textwidth, height = 5.5cm]{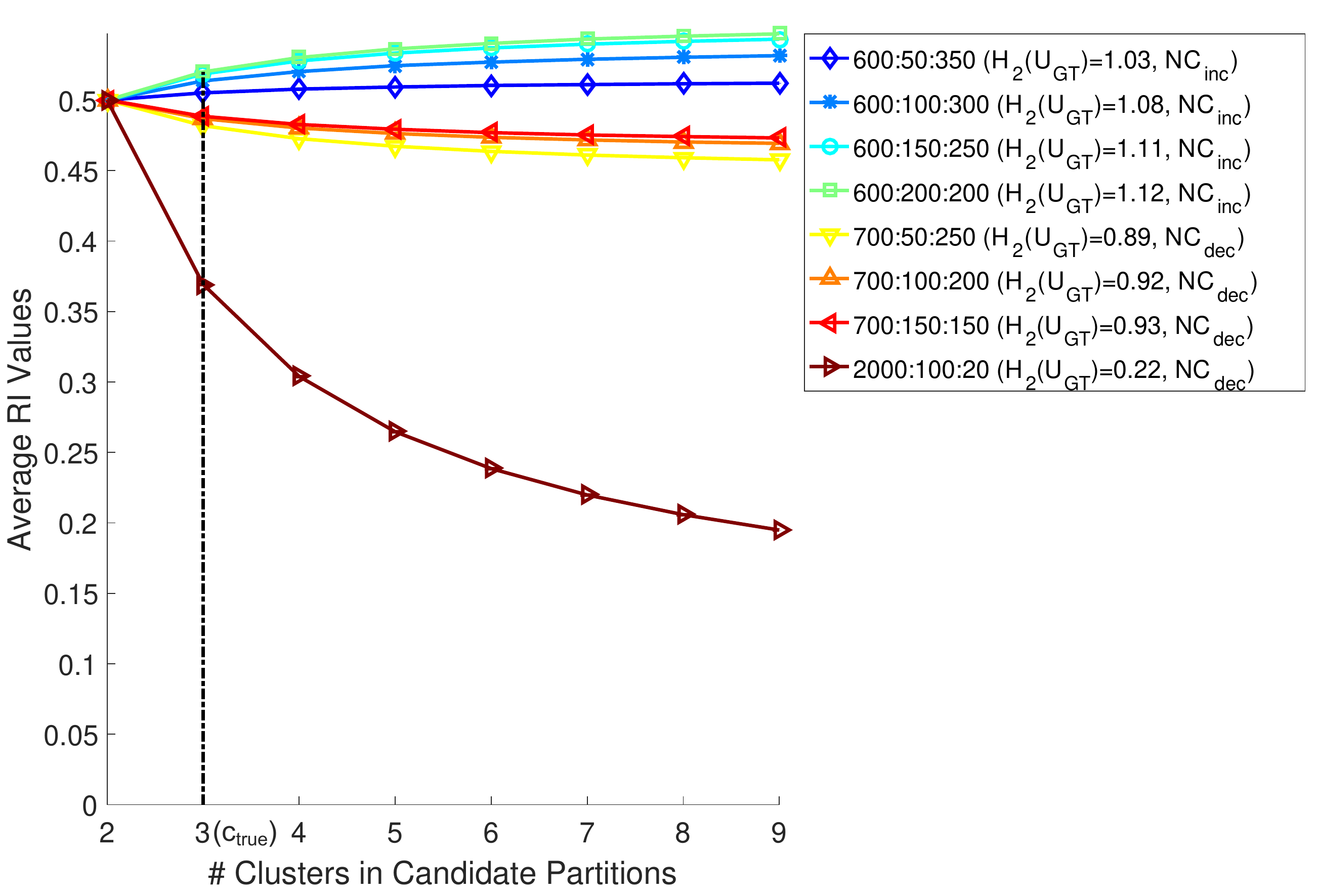}
\caption{$100$ trial average RI values with $c_i$ ranging from $2$ to $9$ for $c_{true}=3$. $n_1:n_2:n_3$ indicates the sizes of the three clusters and the corresponding $H_2(U_{GT})$ values. 
}
\label{fig:1}
\end{figure}

Given a ground truth partition $U_{GT}$, Theorem~\ref{thm:rih2u} provides a test for the NC bias status of the RI. Compute the quadratic entropy $H_2(U_{GT})$ of the reference matrix $U_{GT}$ and compare it to value $1$, and use Theorem~\ref{thm:rih2u} to determine the type of bias. 
Figure~\ref{fig:1} illustrates the relationship between $H_2(U_{GT})$ and $1$ on the Rand index graphically. Figure~\ref{fig:1} is based on the same experimental setting as in Example $2$ but with $N=1000$, and a different distribution $P$ in the ground truth and $c_{true}=3$. Next, we show that we can also judge the NC bias and GT bias of RI by comparing $\sum_{i=1}^r (p_i)^2$ and $\frac{1}{2}$. Next we introduce the Corollary~\ref{cola:pi2} which is the basis for the following theorems. 

\begin{corollary} \label{cola:pi2}
Let $U_{GT} \in M_{hrN}$ be a ground truth partition with $r$ subsets $\{u_1, \ldots, u_r\}$, and let $P= \{p_1, \ldots, p_r\}$ and $p_i = \frac{|u_i|}{N}$. Let $CP = \{V_1, \ldots, V_m\}$ be a set of generated partitions with different numbers of clusters, where $ V_i\in M_{hc_iN}$ contains $c_i$ clusters which are balanced, $2 \leq c_i \leq N$.   Assuming $U_{GT}$ and $V_i \in CP$ are statistically independent, we have
\begin{enumerate}
	\item if $\sum_{i=1}^r (p_i)^2 > \frac{1}{2}$, then $RI$ has NCdec bias;
	\item if $\sum_{i=1}^r (p_i)^2 = \frac{1}{2}$, then $RI$ has NCneu bias;
	\item if $\sum_{i=1}^r (p_i)^2 < \frac{1}{2}$, then $RI$ has NCinc bias.
\end{enumerate}
\end{corollary}

Next, we introduce another theorem which helps us understand how do the prior probabilities $\{p_i\}$ and the number of subsets $r$ in the ground truth $U_{GT}$ influence the NC bias status of RI.
\begin{theorem} \label{thm:prgt}
Let $U_{GT} \in M_{hrN}$ be a ground truth partition with $r$ subsets $\{u_1, \ldots, u_r\}$, and let $P= \{p_1, \ldots, p_r\}$ and $p_i = \frac{|u_i|}{N}$. Let  $P^\prime=\{p_1^\prime,p_2^\prime,\ldots,p_r^\prime\}$ denote $P$ sorted into descending order, where $p_1^\prime \geq p_2^\prime\ldots \geq p_r^\prime$.  Let $CP = \{V_1, \ldots, V_m\}$ be a set of generated partitions with different numbers of clusters, where $ V_i\in M_{hc_iN}$ contains $c_i$ clusters which are balanced, $2 \leq c_i \leq N$.   Assuming $U_{GT}$ and $V_i \in CP$ are statistically independent, then RI has GT bias. In addition, depending on $P$ and $r$, we have: \\
\textbf{When $r  > 2$ }
\begin{enumerate}
\item if $p_1^\prime > \frac{1}{2}$, and \\
%	\begin{itemize}
 if $p_1^\prime (p_1^\prime - \frac{1}{2}) > \sum_{i=2}^r p_i^\prime (\frac{1}{2}-p_i^\prime)$, then RI has NCdec bias;\\
 if $p_1^\prime (p_1^\prime-\frac{1}{2}) = \sum_{i=2}^r p_i^\prime (\frac{1}{2} - p_i^\prime)$, then RI has NCneu bias; \\
 if $p_1^\prime (p_1^\prime-\frac{1}{2}) < \sum_{i=2}^r p_i^\prime (\frac{1}{2} -  p_i^\prime )$, then RI has NCinc bias.
	%\end{itemize}
\item if $p_1^\prime = \frac{1}{2}$, then RI has NCinc bias;
\item if $p_1^\prime < \frac{1}{2}$, then RI has NCinc bias.
\end{enumerate}	
\textbf{When $r=2$} 
\begin{enumerate}
\item if $p_1^\prime > \frac{1}{2}$, then RI has NCdec bias;
\item if $p_1^\prime = \frac{1}{2}$,  then RI has NCneu bias.
\end{enumerate}
\end{theorem}

Theorem~\ref{thm:prgt} tells us how the ground truth distribution $P^\prime$ and the number of clusters $r$ of $U_{GT}$  affect the Rand index and helps us judge the NC bias status based on $P^\prime$ and $r$. For example, if $r>2$, $p_1^\prime > \frac{1}{2}$ and $p_1^\prime(p_1^\prime-\frac{1}{2}) > \sum_{i=2}^r p_i^\prime(\frac{1}{2} - p_i^\prime)$, then RI has NCdec bias (e.g., $r=3$, $p_1^\prime = \frac{2}{3}, p_2^\prime = \frac{1}{4}$ and $p_3^\prime = \frac{1}{12}$). If $r=2$ and $p_1^\prime=\frac{1}{2}$, then RI has NCneu bias. Thus RI has GT bias.

The above discussion and theoretical analysis are in a more general sense. 
Next, we discuss GT1 bias and GT2 bias of the RI, which are two specific types of GT bias with certain conditions imposed on the ground truth. This will also help explain and judge the NC bias behaviours of the indices in the empirical test shown Section~\ref{sec:test}.
\subsubsection{GT1 bias and GT2 bias}
First, we start by introducing a theorem for GT1 bias of RI. 
\begin{theorem} \label{thm:gt1}
Let $U_{GT} \in M_{hrN}$ be a crisp ground truth partition with $r$ balanced subsets $\{u_1, \ldots, u_r\}$, i.e., $p_i = \frac{|u_i|}{N} = \frac{1}{r}$.
 Let $CP = \{V_1, \ldots, V_m\}$ be a set of generated partitions with different numbers of clusters, where $ V_i\in M_{hc_iN}$ contains $c_i$ clusters which are balanced, $2 \leq c_i \leq N$.   Assuming $U_{GT}$ and $V_i \in CP$ are statistically independent, then RI suffers from GT1 bias. More specifically,
\begin{enumerate}
\item if $r=2$, then RI has NCneu bias;
\item if $r>2$, then RI has NCinc bias.
\end{enumerate}	
\end{theorem}
Theorem~\ref{thm:gt1} provides an explanation of how GT1 bias influences RI. For example, it is easier to understand the behaviour of RI shown in the GT1 bias testing in Section~\ref{subsec:gt1t} (Figures~\ref{subfig:ri2c} and~\ref{subfig:ric50}).
Next, we introduce a theorem for the GT2 bias of RI.

\begin{theorem} \label{thm:gt2}
Let $U_{GT} \in M_{hrN}$ be a ground truth partition with $r$ subsets $\{u_1, \ldots, u_r\}$. Assume the first cluster $u_1$ in the ground truth has variable sizes, and the remaining clusters $\{u_1, \ldots, u_r\}$ are uniformly distributed in size across the remaining objects $N - |u_1|$. Let $P= \{p_1, \ldots, p_r\}$ and $p_i = \frac{|u_i|}{N}$, $0<p_i<1$, and $\sum_{i=1}^r p_i =1$. So $p_2=p_3=\ldots=p_r= \frac{1-p_1}{r-1}$.   Let $CP = \{V_1, \ldots, V_m\}$ be a set of generated partitions with different numbers of clusters, where $ V_i\in M_{hc_iN}$ contains $c_i$ clusters which are balanced, $ 1 \leq i \leq m$.   Assuming $U_{GT}$ and $V_i \in CP$ are statistically independent, then RI suffers from GT2 bias. More specifically, let $p^\ast = \frac{2+\sqrt{2(r-1)(r-2)}}{2r}$, we have: \\
\textbf{When $r  > 2$}, 
\begin{enumerate}
\item if $p_1 > p^\ast$, then RI has NCdec bias;
\item if $p_1 = p^\ast$, then RI has NCneu bias;
\item if $p_1 < p^\ast$, then RI has NCinc bias.
\end{enumerate}	
\textbf{When $r=2$} 
\begin{enumerate}
\item if $p_1 = p^\ast$, then RI has NCneu bias;
\item if $p_1 \neq p^\ast$, then RI has NCdec bias.
\end{enumerate}
\end{theorem}

\begin{figure}[b!]
\begin{minipage}[t]{0.6\linewidth}
\centering
\includegraphics[width=\textwidth]{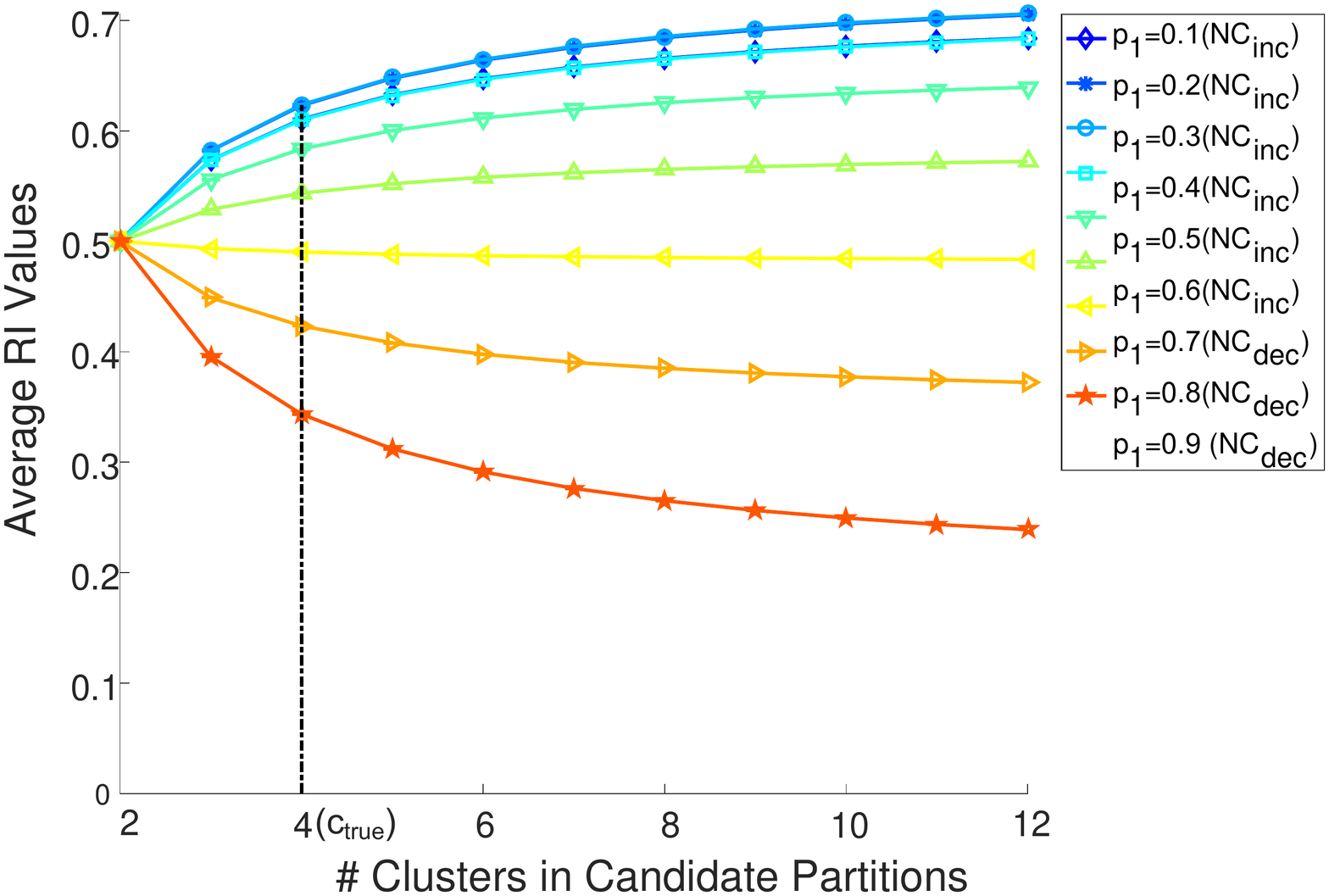}
\caption{$100$ trials average RI values with $c$ in $\{2, \ldots, 12\}$ for $c_{true}=4$. $p_1 = |u_1|/N$, $9$ steps $0.1$ to $0.9$, and the other $3$ clusters uniformly distributed.  When $p_1>p^\ast=0.683$, the $RI$ decreases with $c$ increasing (e.g., $p_1=0.7,0.8,0.9$). When $p_1<p^\ast=0.683$, the $RI$ increases with $c$  (e.g., $p_1 = 0.1,\ldots,0.6$).}
\label{fig:2}
\end{minipage}
\hspace{0.5cm}
\begin{minipage}[t]{0.48\textwidth}
\centering
\includegraphics[width=\textwidth]{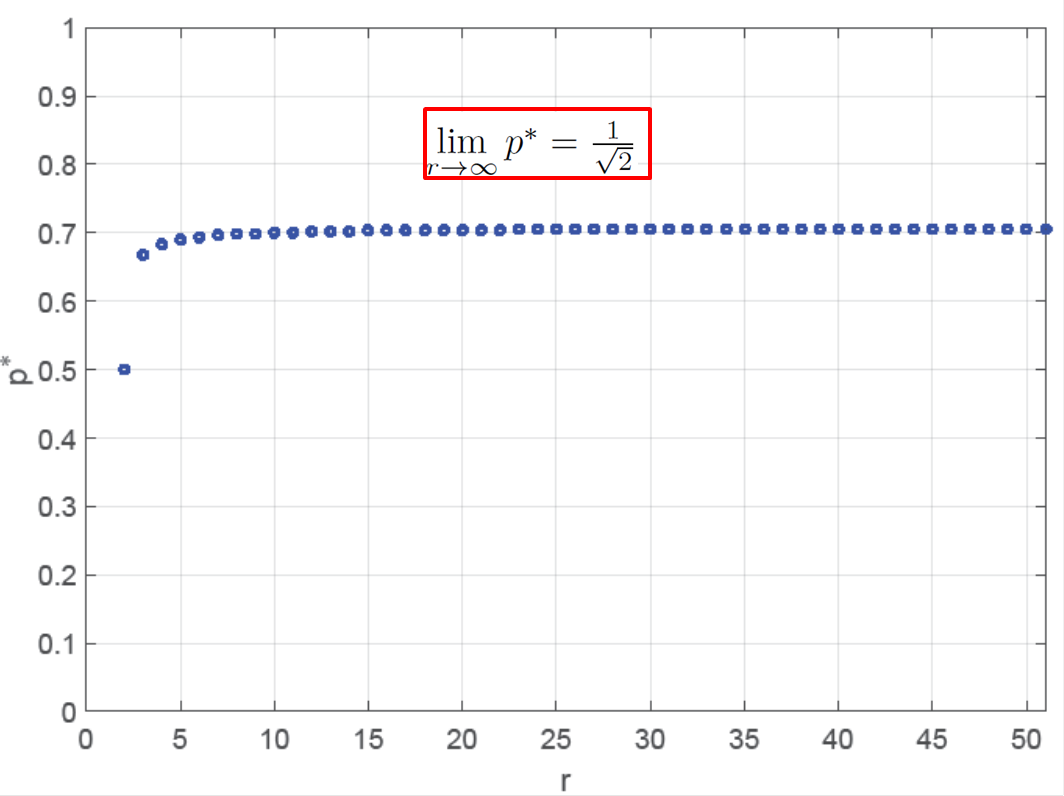}
\caption{The relationship between $p^\ast$ and $r$, for $r$ in $\{2, \ldots, 50\}$.}
\label{fig:3}
\end{minipage}
\end{figure}

Theorem~\ref{thm:gt2} provides an explanation of how GT2 bias affects the RI. For example, in the GT2 bias testing (Section~\ref{subsec:gt2t}), $r=5$, when $p_1 = 0.8 >\frac{1+\sqrt{3}}{4}$ ($\frac{1+\sqrt{3}}{4} \approx 0.683 $), then the $RI$ tends to decreases as $c_i$ increases. Figure~\ref{fig:2} illustrates GT2 bias on  the RI graphically. The basis of this figure is the same experimental setting as Example $2$ in Section~\ref{subsec:exp2} with $N=1000$ and $c_{true}=4$. We also show the relationship between $r$ and $p^\ast$ in Figure~\ref{fig:3} ($r$ takes integer values from $2$ to $50$). Actually, $\lim_{r \rightarrow \infty} p^\ast = \frac{1}{\sqrt{2}}$, where $\frac{1}{\sqrt{2}} \approx 0.7071$.

Next, we conclude our study by giving an experimental example to show that the ARI shows GT bias in certain scenarios.
\section{Example of GT Bias for Adjusted Rand Index (ARI)} \label{sec:ari}
In this section we will illustrate that depending on the set of candidate partitions, ARI can show GT bias behaviour in certain scenarios. Recall that the ARI in Figures~\ref{fig:exUniGT} and~\ref{fig:exSkewGT} had NCneu bias for the method of partition generation used there. We will conduct experiments with a different set of candidates, and will discover that the ARI can be made to exhibit GT bias.
We do two sets of experiments using the following protocols. We first generate ground truth $U_{GT1}$ by randomly choosing $20\%$ of the object labels from $N=100,000$ objects to identify the first cluster. Then, we randomly choose $20\%$ of the object labels from the remaining $80,000$ objects as the second cluster, and finally, we randomly assign the rest of the cluster labels $[3, c_{true}]$ to the remaining objects, where $c_{true} \geq 3$. 
We generate a second ground truth $U_{GT2}$ partition in the following way. We randomly choose $20\%$ of the object labels from $N=100,000$ objects as the first cluster. Then we randomly choose $50\%$ of the object labels from the remaining $80,000$ objects as the second cluster, and finally, we assign the rest of the cluster labels $[3, c_{true}]$ to the rest of objects, where $c_{true} \geq 3$. We set $c_{true}=5$ for both $U_{GT1}$ and $U_{GT2}$.

\begin{figure}[t!]
\centering
\setlength{\fboxsep}{1pt}
\setlength{\fboxrule}{1pt}
\framebox[1\width][c]{
\subfloat[The average ARI values for $U_{GT1}$ and random partitions with different numbers of clusters.]{
\begin{tikzpicture}[squarednode/.style={rectangle, draw=red!60, fill=red!5, very thick, minimum size=5mm},]
\draw(0,0) node[inner sep=0]{\includegraphics[width = 0.4\textwidth]{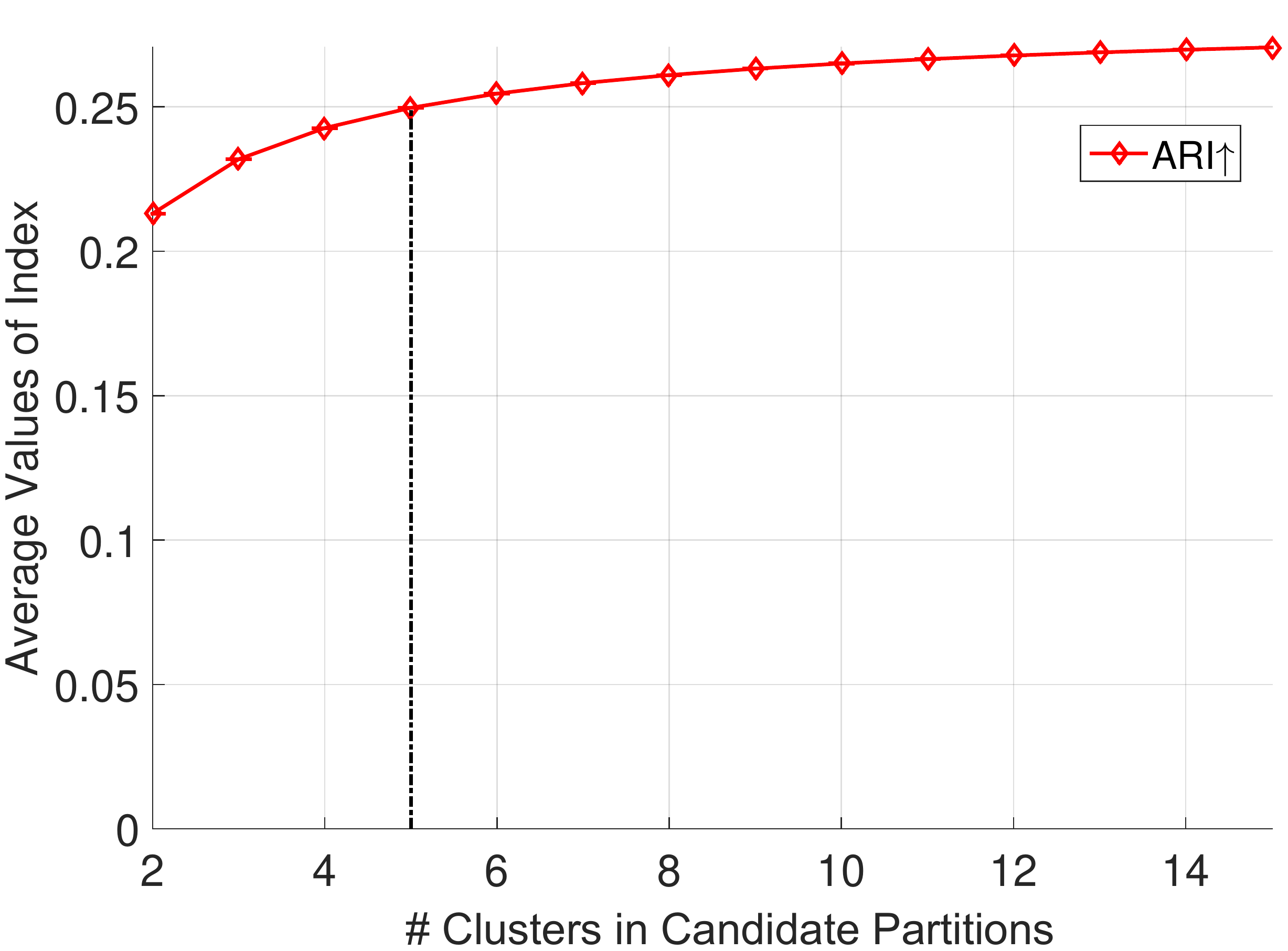}
\label{subfig:ariGT1}};
\draw(1,0) node[squarednode](s-ARI){NCinc};
\end{tikzpicture}
}
\quad
\subfloat[The average ARI values for $U_{GT2}$ and random partitions with different numbers of clusters.]{
\begin{tikzpicture}[squarednode/.style={rectangle, draw=red!60, fill=red!5, very thick, minimum size=5mm},]
\draw(0,0) node[inner sep=0]{\includegraphics[width = 0.4\textwidth]{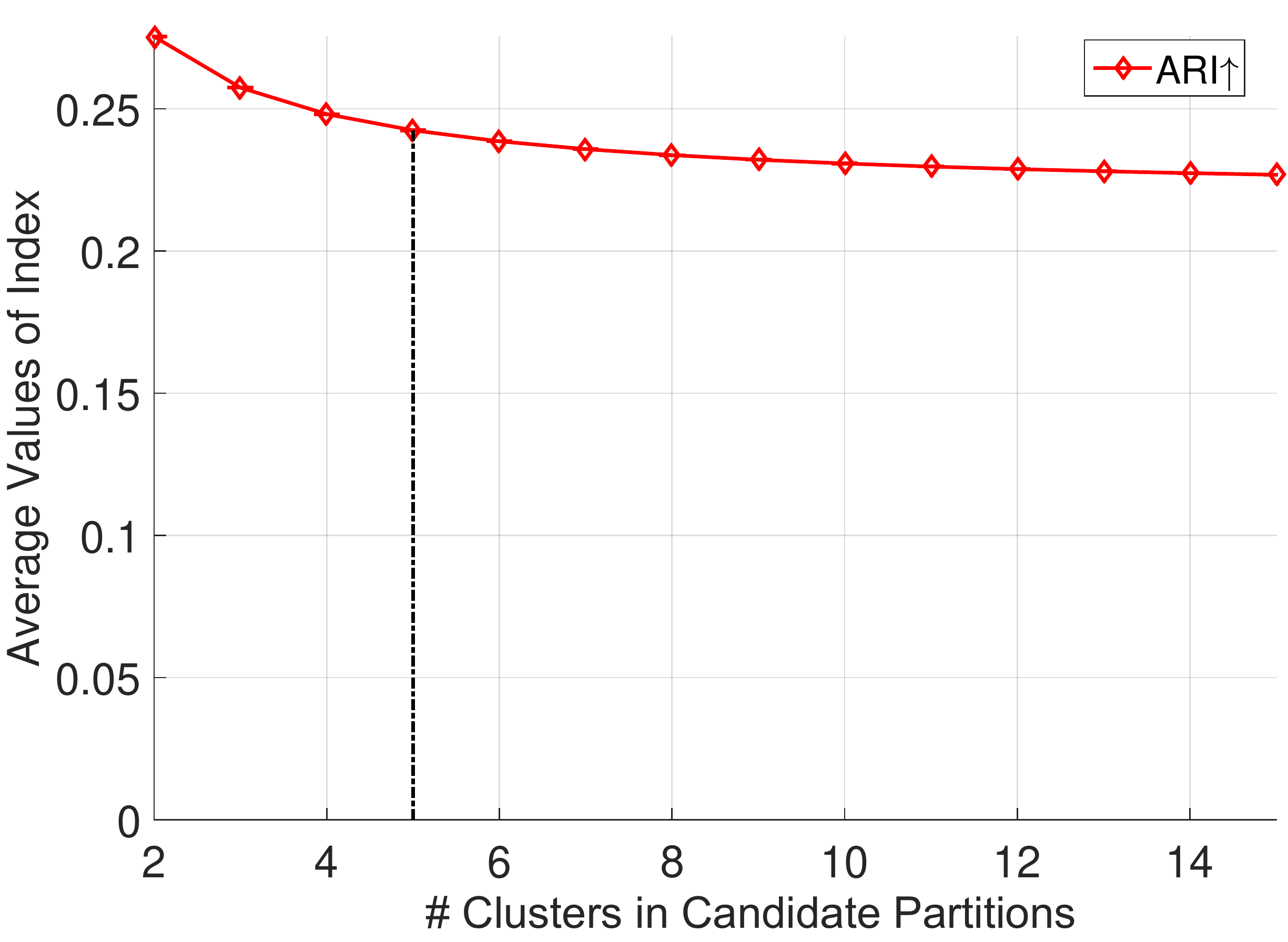}
\label{subfig:ariGT2}};
\draw(0.5,0) node[squarednode](d-ARI){NCdec};
\end{tikzpicture}
}
\tikz[remember picture, overlay] \draw[line width=1pt,-stealth,red] ([xshift=0mm]s-ARI.east) -- ([xshift=0mm]d-ARI.west)node[midway,above,text=black,font=\large\bfseries\sffamily] {$\Rightarrow$GT bias};
}
\caption{$100$ trial average ARI values for two different ground truth and set of candidates with different numbers of clusters. 
%The symbol $\uparrow$ aside the ARI indicates that the ARI is max-optimal.
}
\label{fig:ariGT}
\end{figure}

For these two sets of experiments, we generate $100$ candidate partitions $CP$ in this way. 
For each candidate $V_i \in CP$, we copy the first cluster from $U_{GT1}$ or $U_{GT2}$ as the first cluster in $V_i$. Then, we randomly assign the rest of cluster labels $[2, c_i]$ to the other $80,000$ objects, where $c_i$ ranges from $c=2$ to $c=15$. The results are shown in Figure~\ref{fig:ariGT}. For these two experiments the ARI shows NCinc bias with $U_{GT1}$ and shows NCdec bias with $U_{GT}$. Comparing Figures~\ref{subfig:ariGT1} and~\ref{subfig:ariGT2} shows that for these experiments, the ARI suffers from GT bias. For the exploration of this interesting phenomenon is beyond the scope of this paper, and is an  interesting direction for future work.

\section{Conclusions} \label{sec:cons}
This paper examines several types of bias that may affect external cluster validity indices that are used to evaluate the quality of candidate partitions by comparison with a ground truth partition. They are: 
\begin{inparaenum}[i)]
\item one of two types of NC bias (NCinc, NCdec), which arises when the mathematical model of the external CVI tends to be monotonic in the number of clusters in candidate partitions;
\item GT bias, which arises when the ground truth partitions alters the NC bias status of an external CVI; 
\item GT1 bias, which arises when the numbers of clusters in the ground truth partitions alters the NC bias status of an external CVI; 
\item GT2 bias, which arises when the distribution of the ground truth partitions alters the NC bias status of an external CVI.
\end{inparaenum}

Numerical experiments with $26$ pair-counting based external CVIs established that for the method described in the examples, $5$ of the $26$ suffer from GT1bias and/or GT2bias, viz., the indices due to Rand ($\#1$), Mirkin ($\#3$), Hubert ($\#5$), Gower and Legendre ($\#24$) and Rogers and Tanimoto ($\#25$), the numbers referring to rows in Table~\ref{tb:RIformu}. Actually, the $4$ indices, Mirkin ($\#3$), Hubert ($\#5$), Gower and Legendre ($\#24$) and Rogers and Tanimoto ($\#25$), are all functions of RI. 
We point out that the observed bias behaviour (NC bias, GT1 bias and GT2 bias) of the tested 26 indices was based on a particular way to obtain candidate partitions.
In our experiments the ``clustering algorithm" used to generate the CPs was random draws from $M_{hc_iN}$. It is entirely possible that sets of CPs secured, for example, by running clustering clustering algorithms on a dataset will NOT exhibit the same bias tendencies. This is just another difficulty of external cluster validity indices, as was illustrated by the fact that we could change the bias status of the ARI by changing the method of securing the candidate. 
The major point of this work is to draw attention to the fact that there can be a GT bias problem for external CVIs. 
%The two types of GTbias were illustrated graphically.
% using Monte Carlo simulations for the ground truth and candidates.

We then formulated an explanation for both types of GT bias with Rand Index based on the the Havrda-Charvat quadratic entropy. Our theory explained how RI's NC bias behaviour is influenced by the distribution of the ground truth partition and also the number of clusters in the ground truth. Our major results in Theorem~\ref{thm:rih2u}, which provides a computable test that predicts the NC bias behaviour of the Rand Index, and hence, all external CVIs related to it. Rand Index has been one of the most popular external CVIs due to its simple, natural interpretation and has recently been applied in many research work~\citep{RI2010nature, RI2011a, RI2012b, RI2013a, RI2014a, rivi2015}. Thus, the identified GT bias behaviour for RI with correponding explaination could be helpful for users who apply RI in their work.
Finally, we gave an experimental example showing that the ARI can suffer from GT bias in certain scenarios.  

We believe this to be the first systematic study of the effects of ground truth on the NC bias behaviour of external cluster validity indices. We have termed this GT bias. There are many other external CVIs which have not been tested numerically or analyzed theoretically for GT bias. Our next undertaking will be to study this phenomenon in the more general setting afforded by non pair-counting based external CVIs.

\appendix

\section{Proofs} \label{apdx}
\paragraph{\textbf{Proof of Lemma~\ref{lema:vi2ri}}}
%\begin{proof}
As $U$ and $V$ are statistically independent, we can substitute Equation~\ref{eq:h2uv} into Equation~\ref{eq:vi2}, obtaining
\begin{align}
	VI_2(U,V) & = 2H_2(U,V) - H_2(U) - H_2(V) \nonumber \\
			  & = 2 \big(H_2(U)+H_2(V)-\frac{1}{2}H_2(U)H_2(V)\big) - H_2(U) - H_2(V) \nonumber \\
	        %  & = H_2(U) + H_2(V) -H_2(U)H_2(V) \nonumber \\
	          & = H_2(U) + (1 - H_2(U))H_2(V) 
\end{align}
%\end{proof}
\paragraph{\textbf{Proof of Theorem~\ref{thm:rih2u}}}
%\begin{proof}
According to lemma~\ref{lema:vi2ri}, 
\begin{equation}
VI_2(U_{GT},V) = H_2(U_{GT}) + (1 - H_2(U_{GT}))H_2(V) = a + bx \label{eq:vi2line}
\end{equation}
where $a = H_2(U_{GT})$ and $b = (1-H_2(U_{GT})) = (1-a)$ and $x = H_2(V)$. As any $V_i \in CP$ is uniformly distributed (balanced), then $H_2(V_i) = 2(1-\sum_{i=1}^{c_i} (\frac{1}{c_i})^2)$ (refer to equation~\ref{eq:h2u}) and $H_2(V_i)$ increases as $c_i$ increases. It is clear from equation~(\ref{eq:vi2line}) that for fixed $U_{GT}$, $VI_2$ can be regarded as a straight line with $y$ intercept $a = H_2(U_{GT})$ and slope $b=1-H_2(U_{GT})=(1-a)$, so the rate of growth (or decrease, or neither (flat)) of $VI_2$ depends on $b$. In other words, $VI_2$ could be increasing, decreasing or flat as $c_i$ increases. More specifically, 
\begin{inparaenum}[i)]
\item if $b>0$, then $H_2(U_{GT}) < 1$, thus $VI_2$ increases as $x$ (and $c_i$) increases;
\item if $b=1$, then $H_2(U_{GT})=1$, thus $VI_2$ is constant as $x$ (and $c_i$) increases;
\item if $b<0$, then $H_2(U_{GT}) > 1$, thus $VI_2$ decreases as $x$ (and $c_i$) increases.
\end{inparaenum}
According to Equation~\ref{eq:vi2ri}, we know that $VI_2$ and RI are inversely related. Thus, it is straightforward to prove the statements.

\paragraph{\textbf{Proof of Corollary~\ref{cola:pi2}}}
According to Theorem~\ref{thm:rih2u}, we know that depending on the relationship between $H_2(U_{GT})$ and $1$, i.e., the slope $b$ in Equation~\ref{eq:vi2line}, that RI shows different NC bias status. 
As $H_2(U_{GT}) = 2(1-\sum_{i=1}^r p_i^2)$ (Equation~\ref{eq:h2u}), then $b = 1-H_2(U_{GT})  = 1-2(1-\sum_{i=1}^r p_i^2) = 2(\sum_{i=1}^r p_i^2 - \frac{1}{2})$.
Thus, we know that the slope $b$, i.e., the relationship between $H_2(U_{GT})$ and $1$, depends on the relationship between $\sum_{i=1}^r p_i^2$ and $\frac{1}{2}$. So, the three assertions of the corollary follow by noting the relationship between $\sum_{i=1}^r p_i^2$ and $\frac{1}{2}$.

\paragraph{\textbf{Proof of Theorem~\ref{thm:prgt}}}
%\begin{proof}
According to Corollary~\ref{cola:pi2}, we know that the relationship between $\sum_{i=1}^r p_i^2$ and $\frac{1}{2}$ influences the NC bias status of RI.  It is straightforward to see that $p_i$ and $r$ influence the relationship between  $\sum_{i=1}^r p_i^2$ and $\frac{1}{2}$,  thus $p_i$ and $r$ can potentially alter the NC bias status of RI. We have
\begin{equation}
	\sum_{i=1}^r (p_i^\prime)^2 - \frac{1}{2} = \sum_{i=1}^r (p_i^\prime)^2 - \frac{1}{2}\sum_{i=1}^r p_i^\prime 
  %  & = \sum_{i=1}^r p_i^\prime (p_i^\prime-\frac{1}{2}) \nonumber \\
	 = p_1^\prime(p_1^\prime-\frac{1}{2}) + \sum_{i=2}^r p_i^\prime (p_i^\prime - \frac{1}{2}) \label{eq:pihalf}
\end{equation}

Please note that $p_1^\prime$ is the biggest cluster's density in the ground truth, based on which we discuss and summarize the influence of $p_i$ and $r$ on the NC bias status of RI. We can discuss the relationship between $p_1^\prime(p_1^\prime-\frac{1}{2})$ and $\sum_{i=2}^r p_i^\prime (p_i^\prime - \frac{1}{2})$, which is equivalent to  the relationship between $\sum_{i=1}^r p_i^2$ and $\frac{1}{2}$, for the different NC bias status.\\
\textbf{When $r  > 2$}:
\begin{inparaenum}[1)]
\item if $p_1^\prime > \frac{1}{2}$, because $\sum_{i=1}^r p_i^\prime=1$, then $p_2^\prime, \ldots, p_r^\prime < \frac{1}{2}$. Thus, with the help of Corollary~\ref{cola:pi2}, we have:
\begin{inparaenum}[i)]
\item if $p_1^\prime (p_1^\prime - \frac{1}{2}) > \sum_{i=2}^r p_i^\prime (\frac{1}{2}-p_i^\prime)$, then $\sum_{i=1}^r (p_i^\prime)^2 > \frac{1}{2}$, thus RI has NCdec bias;
\item if $p_1^\prime (p_1^\prime-\frac{1}{2}) = \sum_{i=2}^r p_i^\prime (\frac{1}{2} - p_i^\prime)$, then $\sum_{i=1}^r (p_i^\prime)^2 = \frac{1}{2}$, thus RI has NCneu bias;
\item if $p_1^\prime (p_1^\prime-\frac{1}{2}) < \sum_{i=2}^r p_i^\prime (\frac{1}{2} -  p_i^\prime )$, then $\sum_{i=1}^r (p_i^\prime)^2 < \frac{1}{2}$, thus RI has NCinc bias.
\end{inparaenum}\\
\item if $p_1^\prime = \frac{1}{2}$, then $p_2^\prime, \ldots, p_r^\prime < \frac{1}{2}$ and $\sum_{i=1}^r (p_i^\prime)^2 < \frac{1}{2}$, thus RI has NCinc bias;  
\item if $p_1^\prime < \frac{1}{2}$, then $p_2^\prime, \ldots, p_r^\prime < \frac{1}{2}$ and $\sum_{i=1}^r (p_i^\prime)^2 < \frac{1}{2}$, thus RI has NCinc bias.
\end{inparaenum}\\	
\textbf{When $r=2$}: 
\begin{inparaenum}[1)]
\item if $p_1^\prime > \frac{1}{2}$, then $p_1^\prime (p_1^\prime - \frac{1}{2}) > p_2^\prime (\frac{1}{2}-p_2^\prime) $ and $\sum_{i=1}^r (p_i^\prime)^2 > \frac{1}{2}$, thus RI has NCdec bias;
\item if $p_1^\prime = \frac{1}{2}$, then $p_2^\prime = \frac{1}{2}$ and $\sum_{i=1}^r (p_i^\prime)^2 = \frac{1}{2}$, thus RI has NCneu bias.
\end{inparaenum}
Thus, the RI suffers from GT bias according to the distribution of ground truth $P$ and the number of clusters $r$ in the ground truth.
%\end{proof}

\paragraph{\textbf{Proof of Theorem~\ref{thm:gt1}}}
Corollary~\ref{cola:pi2} shows that the NC bias of the RI depends on the relationship between $\sum_{i=1}^r(p_i)^2$ and $\frac{1}{2}$. Since $p_i = \frac{1}{r}$, then $\sum_{i=1}^r(p_i)^2 - \frac{1}{2} = \frac{1}{r}-\frac{1}{2}$.
Then, according to Corollary~\ref{cola:pi2}, we have 
\begin{inparaenum}[i)]
\item if $r=2$, then RI has NCneu bias;
\item if $r>2$, then RI has NCinc bias.
\end{inparaenum}
By definition~\ref{def:gt1bias}, different values for $r$ in $U_{GT}$, i.e., $r=2$ or $r>2$, result in different NC bias status for the RI, thus RI has GT1 bias.

\paragraph{\textbf{Proof of Theorem~\ref{thm:gt2}}}
According to Corollary~\ref{cola:pi2}, we know that the relationship between $\sum_{i=1}^r(p_i)^2$ and $\frac{1}{2}$ determines the NC bias status of the RI. 
As $p_i = \frac{1-p_1}{r-1}$, $i = 2, \ldots, r$, we have:
\begin{align}
  \sum_{i=1}^r p_i^2 - \frac{1}{2} & = p_1^2 + \sum_{i=2}^r p_i^2 - \frac{1}{2}\nonumber \\
           % & =   p_1^2 + (r-1)\big(\frac{1-p_1}{r-1}\big)^2 - \frac{1}{2} \nonumber \\ 
            & = \frac{r}{r-1}p_1^2 - \frac{2}{r-1}p_1 + \frac{3-r}{2(r-1)} \label{eq:rp1}
\end{align}
Equation~\ref{eq:rp1} is quadratic in $p_1$, and has one real positive root $p^\ast = \frac{2+\sqrt{2(r-1)(r-2)}}{2r}$ in our case. Then:\\ 
\textbf{When $r  > 2$}:
\begin{inparaenum}[i)]
\item if $p_1 > p^\ast$, then $\sum_{i=1}^r (p_i)^2 > \frac{1}{2}$, thus RI has NCdec bias;
\item if $p_1 = p^\ast$, then $\sum_{i=1}^r (p_i)^2 = \frac{1}{2}$, thus RI has NCneu bias;
\item if $p_1 < p^\ast$, then $\sum_{i=1}^r (p_i)^2 < \frac{1}{2}$, thus RI has NCinc bias.
\end{inparaenum}\\
\textbf{When $r=2$}:
\begin{inparaenum}[i)]
\item if $p_1 = p^\ast$, then $\sum_{i=1}^r (p_i)^2 = \frac{1}{2}$, thus RI has NCneu bias;
\item if $p_1 \neq p^\ast$, then $\sum_{i=1}^r (p_i)^2 > \frac{1}{2}$, thus RI has NCdec bias.
\end{inparaenum}

\section*{Acknowledgement}
This work is supported by the Australian Research Council via grant numbers *** and ***.
\bibliography{ribiasRef}

\end{document}